\documentclass[11pt]{article}

\usepackage[margin=1in]{geometry}
\usepackage{times}
\usepackage{natbib}
\usepackage{authblk}

\usepackage{amsmath,amsfonts,bm}

\def\eqref#1{equation~\ref{#1}}

\def\1{\bm{1}}

\DeclareMathAlphabet{\mathsfit}{\encodingdefault}{\sfdefault}{m}{sl}
\SetMathAlphabet{\mathsfit}{bold}{\encodingdefault}{\sfdefault}{bx}{n}

\usepackage{graphicx}
\usepackage{hyperref}
\usepackage{url}
\usepackage{booktabs}
\usepackage{tabularx}
\usepackage[table]{xcolor}

\definecolor{linkblue}{HTML}{355C7D}
\hypersetup{
  colorlinks=true,
  linkcolor=linkblue,
  citecolor=linkblue,
  urlcolor=linkblue
}

\title{\textbf{Beyond Correctness: Compute-Responsive Data Selection for Reasoning Model Post-Training}}

\author{
Liang He \quad
Xiangyu Li \quad
Yuxin Wang \quad
Hanbin Zhang \quad
Mingyu Cao \\
Zhaomiao Guo \quad
Xilu Wang
\\
{\small
University of Texas at Austin
\quad
Advanced Micro Devices
\quad
University of Warwick
\quad
University of Surrey
}
}

\date{}

\begin{document}

\maketitle

\begin{abstract}
Reasoning language models increasingly rely on test-time compute scaling, yet
most existing work studies how to allocate inference computation while treating
the model itself as fixed.
We ask a complementary question: can reasoning models be post-trained to make
better use of a given amount of test-time compute, and which examples are most
useful for this goal?
Our key observation is that reasoning problems differ substantially in their
response to additional computation.
Some are already solved under limited compute, others remain unresolved even
after further reasoning, while a distinct subset changes from complete initial
failure to success when additional reasoning is provided.
We call these \emph{compute-responsive} (CR) examples and introduce
\emph{Compute-Responsive Data Selection} (CRDS), a post-training data-selection
principle that prioritizes examples with empirically demonstrated marginal
utility from additional reasoning.
On Qwen3-8B, SFT on CR examples recovers 45.3\% of held-out low-compute
failures, compared with 40.6\% when training on already-solved examples.
The advantage extends beyond the selected CR split: SFT-CR reaches 51.8\%
overall accuracy on an independently sampled test distribution, with the gain
concentrated on compute-responsive problems.
CR training also shifts the compute--accuracy frontier across inference budgets
from 512 to 4,096 tokens, outperforming matched non-CR training throughout and
reaching 48.1\% recovery at the largest budget.
The same pattern is supported across multiple data sources, remains visible
under on-policy distillation, and extends to Llama-3.1-8B-Instruct when CR sets
are reconstructed separately for that model.
Better test-time compute utilization can therefore be learned through
post-training, with recoverable responses to additional reasoning providing a
particularly informative signal for selecting training examples.
\end{abstract}

\section{Introduction}
\label{sec:intro}

Test-time computation has emerged as an important scaling dimension for
reasoning language models.
Modern systems can improve predictions by spending additional inference-time
computation on longer derivations, repeated attempts, verification, search, or
refinement\citep{wei2022chain,wang2023selfconsistency,lightman2024verify,yao2023tree,madaan2023selfrefine}.
This has motivated substantial work on test-time scaling and adaptive
inference, where a central question is how much computation should be allocated
to each problem~\citep{snell2024scaling,wen2026not}.

Most existing approaches, however, treat the reasoning model as a fixed
consumer of computation: once additional compute is allocated, it is assumed
that the model can productively convert it into better reasoning. Our
experiments challenge this assumption. Additional realized generation can fail
to produce additional correct solutions, suggesting an important distinction:
available compute is not necessarily useful compute. This motivates a
complementary question:

\begin{center}
\textbf{\emph{Can reasoning models be post-trained to make better use of
test-time compute, and if so, which training examples are most useful for this
goal?}}
\end{center}

\begin{figure}[t]
\centering
\includegraphics[width=\linewidth]{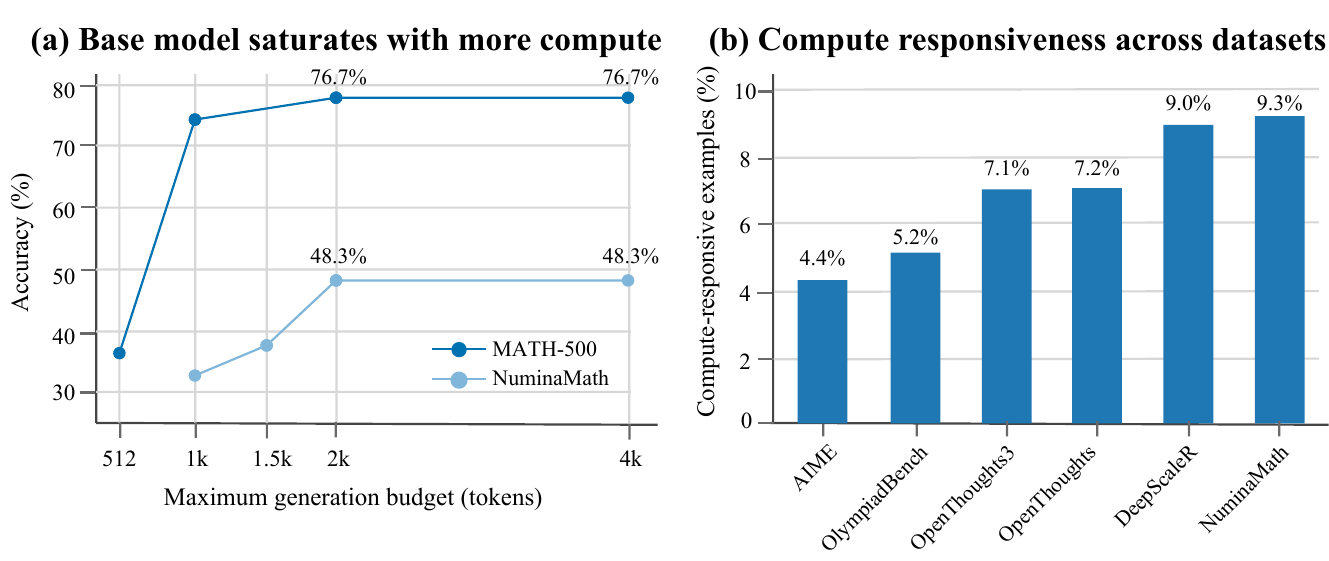}
\caption{
\textbf{Heterogeneous response to additional test-time computation.}
(a) Increasing the generation budget improves base-model accuracy at lower
budgets, but beyond a certain budget, additional computation no longer
translates into higher accuracy.
(b) Compute-responsive examples occur at substantially different rates across
reasoning datasets.
}
\label{fig:motivation}
\end{figure}

Figure~\ref{fig:motivation} summarizes the two observations that motivate our
approach. 
In Figure~\ref{fig:motivation}(a), increasing the generation budget improves
base-model accuracy in lower-budget comparisons, but the marginal gain
diminishes and eventually disappears in the highest-budget comparison shown.
In Figure~\ref{fig:motivation}(b), responsiveness to additional reasoning varies
considerably across datasets. Some problems are already solved under limited
compute, others remain unresolved, while a distinct subset transitions from a
complete initial failure to a correct solution after additional reasoning.
We term these \emph{compute-responsive} (CR) problems.
CR problems provide a particularly informative form of post-training
supervision. Already-solved examples demonstrate behavior the model can already
execute under limited compute, while unresolved examples provide no successful
reasoning path under the available additional computation. CR examples lie
between these two cases: they reveal problems for which a successful reasoning
path is accessible to the current model but is not reliably reached in its
initial attempt. This suggests that post-training data should not be treated as
equally useful for improving compute utilization~\citep{zhou2023lima,wang2023selfinstruct}.

Based on this observation, we introduce \emph{Compute-Responsive Data
Selection} (CRDS), a post-training data-selection principle that prioritizes
examples according to whether additional reasoning actually changes the
outcome. The goal is not simply to train on harder or previously incorrect
problems, but to identify failures for which additional reasoning has already
demonstrated recoverability.
Constructing genuine CR data requires more than increasing the maximum
generation length. Our early experiments showed that larger token budgets often
capture only \emph{completion responsiveness}, where a truncated trajectory is
allowed to finish, rather than a genuine change in reasoning. We also found
that deterministic revision can reproduce the original error and that brittle
answer matching can create formatting-induced false recoveries. We therefore
restrict CR candidates to complete initial failures, probe recoverability
through sampled blind re-solving and audit-based revision, and normalize
answers before assigning CR labels.

We primarily study CRDS under supervised fine-tuning, where its effect can be
isolated through matched training sets that differ only in how examples are
selected. We additionally apply the same selection principle to on-policy
distillation, where supervision follows states visited by the evolving student
policy, providing a complementary test of whether the CR signal remains useful
beyond fixed offline supervision. CRDS changes only the post-training
distribution: at inference time, the trained model receives only the original
problem, without previous failures, CR labels, repair prompts, or
collection-time verifiers.
Using Qwen3-8B, we find that post-training itself already improves recovery of
low-compute failures, but the choice of training data matters substantially.
CR-selected SFT provides the strongest recovery among matched data-selection
baselines and produces a substantially better compute--accuracy frontier across
budgets from 512 to 4,096 tokens. The improvement persists on an independently
sampled test distribution, is supported across multiple data sources, and also
appears under on-policy distillation. We further test the same selection
principle across additional model families and scales.
Our main contributions are:
\begin{itemize}
\item We distinguish \emph{compute allocation} from
\emph{compute utilization} and show that additional test-time computation does
not necessarily translate into better reasoning.


\item We introduce \emph{Compute-Responsive Data Selection} (CRDS), a
post-training data-selection principle identifying examples where
additional reasoning converts complete initial failure into success, while
separating genuine reasoning recovery from truncation and evaluation artifacts.

\item We show that post-training data selection affects compute
utilization: CR-selected supervision outperforms matched random and
already-solved controls, shifts the compute--accuracy frontier, and remains
useful across datasets, objectives, and model families.
\end{itemize}

\section{Related Work}
\label{sec:related}

\paragraph{Test-time compute scaling.}
Recent reasoning systems improve inference by increasing test-time computation
through longer generations, repeated sampling, decomposition, search,
verification, tool use, or iterative refinement
\citep{zhou2023leasttomost,gao2023pal,chen2023program,yao2023react,shinn2023reflexion}.
These approaches establish inference computation as an important scaling axis,
but primarily study how a fixed model should spend additional computation.
Our work asks a prior question: for which problems does additional reasoning
actually yield marginal utility?---because only by identifying such problems
can we discuss how to allocate computation to them, and this
``responsiveness'' itself also becomes a viable principle for selecting
post-training examples.

\paragraph{Adaptive computation and routing.}
A related literature allocates different amounts of computation to different
examples according to difficulty, confidence, uncertainty, expected
improvement, or task utility
\citep{liu2020fastbert,schuster2022calm,schwartz2020righttool,chen2024frugalgpt,ong2025routellm}.
These methods optimize \emph{where} a limited inference budget should be
spent. CRDS operates at a different layer: it modifies the model so that
computation allocated can be used more productively. The two perspectives can therefore be combined rather than viewed as alternatives.

\paragraph{Reasoning post-training and distillation.}
Supervised fine-tuning, reinforcement learning, preference optimization,
knowledge distillation, and on-policy distillation have all been used to
improve reasoning models
\citep{rafailov2023dpo,shao2024deepseekmath,hsieh2023distilling,agarwal2024onpolicy}.
Although these methods differ in their optimization objectives, all depend
critically on which examples provide supervision.
Conventional selection criteria emphasize correctness, difficulty, quality,
diversity, or teacher confidence.
Compute responsiveness is different: an example is selected because additional
reasoning has empirically demonstrated marginal utility for the current model.

\paragraph{Self-correction and revision.}
Prior work has also studied critique, reflection, verification, and
self-correction~\citep{bai2022constitutional,welleck2023selfcorrect,gou2024critic}.
Our use of revision differs in purpose.
Blind re-solving and audit-based revision are primarily \emph{probes} used to
determine whether an initially failed problem lies within the model's
recoverable reasoning boundary.
The central contribution is therefore not a new revision prompt, but the
post-training signal obtained from the observed response to additional
reasoning.

\section{Compute Responsiveness as a Selection Principle}
\label{sec:method}

\begin{figure}[t]
\centering
\includegraphics[width=1\linewidth]{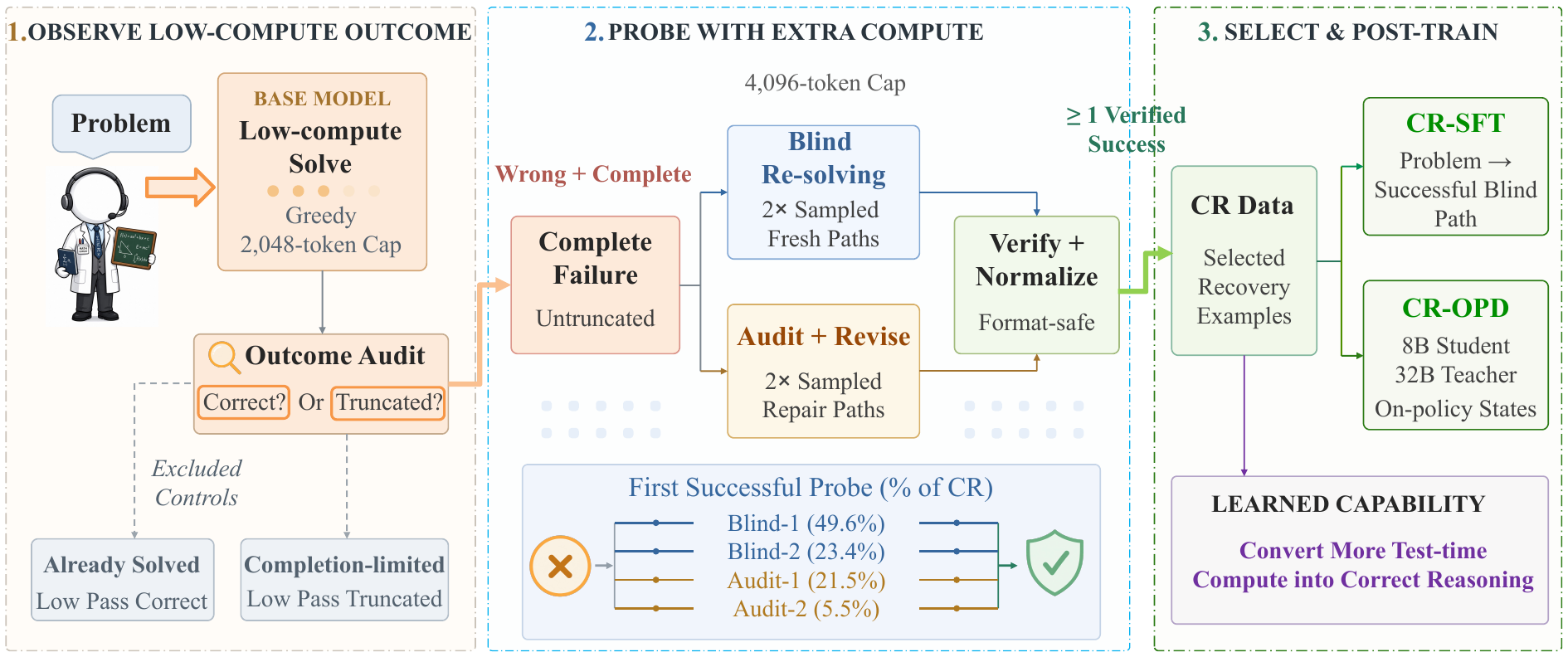}
\caption{
\textbf{Compute-responsive data selection.}
Starting from complete initial failures, additional reasoning is used to
identify problems for which at least one follow-up trajectory succeeds.
The resulting compute-responsive examples are selected for post-training;
no collection-time trajectory or verifier is required at inference.
}
\label{fig:method}
\end{figure}

Our method has two stages.
We first identify problems whose outcomes improve under additional reasoning,
and then use these behavioral differences to select post-training data.
Figure~\ref{fig:method} summarizes the framework.

\paragraph{Compute Responsiveness}

For a problem $x_i$, let the base model produce an initial reasoning trajectory
under a fixed inference procedure,
\begin{equation}
y_i^{\mathrm{low}}
\sim
\pi_0(\cdot \mid x_i,c_{\mathrm{low}}),
\end{equation}
and let an external verifier determine correctness,
\begin{equation}
s_i^{\mathrm{low}}
=
\mathrm{Eval}(y_i^{\mathrm{low}})
\in \{0,1\}.
\end{equation}
We also record whether the generation is truncated by the maximum generation
budget, denoted by $t_i^{\mathrm{low}}=1$.
A CR problem must be a \emph{complete initial failure}, rather than
an unfinished generation:
\begin{equation}
s_i^{\mathrm{low}}=0,
\qquad
t_i^{\mathrm{low}}=0.
\label{eq:low_candidate}
\end{equation}
This excludes cases in which increasing the token cap merely allows a
truncated trajectory to finish, which we refer to as
\emph{completion responsiveness}.
For each complete failure, we generate additional reasoning trajectories using
sampled blind re-solving and audit-based revision.
Let $\mathcal{H}_i$ denote the resulting set of additional trajectories.
We define
\begin{equation}
x_i \in D_{\mathrm{CR}}
\iff
s_i^{\mathrm{low}}=0,\;
t_i^{\mathrm{low}}=0,\;
\exists y \in \mathcal{H}_i :
\mathrm{Eval}(y)=1.
\label{eq:cr_definition}
\end{equation}
Answers are normalized before applying
Equation~\ref{eq:cr_definition} so that formatting-only changes are not counted
as reasoning recovery.
CR is therefore defined by observed behavior under a fixed model and
additional-reasoning protocol.
A problem may be already solved by one model, compute-responsive for another,
or remain unresolved under the available reasoning procedure.
This separates responsiveness from intrinsic problem difficulty.

\paragraph{Constructing the CR Training Set}

Constructing genuine CR examples requires more than simply increasing
\texttt{max\_new\_tokens}.
Pilot experiments revealed three common confounds: completion effects from
truncated low-budget trajectories, repeated errors under deterministic
revision, and false recoveries caused by brittle answer matching.
We therefore restrict candidates to complete initial failures, probe each
candidate using multiple sampled blind and audit trajectories, and normalize
answers before assigning CR labels.
The evolution of this collection pipeline is described in
Appendix~\ref{app:pipeline_evolution}, with representative normalization
artifacts reported in Appendix~\ref{app:format_artifacts}.
During large-scale collection, CR rates vary across datasets, from 4.4\% on
AIME and 5.2\% on OlympiadBench to 9.0\% on DeepScaleR and 9.3\% on the
processed NuminaMath-CoT subset.
Although our collection protocol allows up to four additional trajectories per
complete failure, most CR examples are detected much earlier.
As shown in Appendix~\ref{app:first_hit_cost}, a single blind re-solving attempt
recovers 49.6\% of all CR examples, and two blind attempts recover 73.0\%.
The average first-hit cost among CR examples is only 1.83 additional
generations, indicating that the four-attempt protocol is mainly used to
improve recall rather than being the typical discovery cost.
At the same time, audit-based revision accounts for 27.0\% of CR examples whose
first success is not found by blind re-solving, showing that the blind and
audit probes are not redundant.
After validation, normalization, and problem-level deduplication, the final CR
corpus contains 1,065 unique problems, split into 853 training, 106 validation,
and 106 test examples.
All trajectories associated with the same underlying problem remain in the same
split.
Full collection statistics, source composition, and the data schema are
reported in Appendices~\ref{app:collection_statistics},
\ref{app:split_statistics}, and~\ref{app:data_schema}.

\paragraph{Compute-Responsive Supervised Fine-Tuning}

Our primary instantiation uses ordinary supervised fine-tuning so that the
effect of data selection can be isolated directly.
For each CR training problem with a validated successful blind trajectory, the
input contains only the original problem and the target is the successful
reasoning trajectory; the failed low-compute output is not included.
This design ensures that any inference-time improvement must come from the
post-trained model itself rather than from conditioning on previous failures,
repair prompts, CR labels, or collection-time verifiers.
We compare CR-selected supervision against matched alternatives that use the
same number of examples and the same training procedure but differ in where the
targets come from: incorrect low trajectories from CR problems, random correct
trajectories, correct trajectories from already-solved problems, and successful
trajectories from CR problems.
This controlled design isolates whether compute responsiveness identifies a
more useful region of the training distribution, rather than merely whether
additional fine-tuning is beneficial.
\paragraph{Compute-Responsive On-Policy Distillation}

We also apply the same data-selection principle to on-policy distillation
(OPD) as a complementary training setting.
Unlike SFT, which optimizes against fixed offline targets, OPD samples
trajectories from the current student policy and obtains teacher supervision on
states the student actually visits.
The supervised state distribution therefore changes as the student changes.
We use this setting to test whether compute responsiveness remains informative
when the learning objective already adapts to the evolving student policy.
The OPD variants differ primarily in which problems are selected for training:
all available problems, already-solved problems, or CR problems.
\section{Experimental Setup}
\label{sec:setup}

\paragraph{Models, data, and CR construction.}
Our primary experiments use Qwen3-8B~\citep{qwen3technicalreport} with
\texttt{enable\_thinking=False}.
The initial CR collection pass uses greedy decoding with a maximum generation
budget of 2,048 tokens.
For each complete initial failure, recoverability is probed using sampled blind
re-solving and audit-based revision.
The CR corpus is constructed from six mathematical reasoning sources:
NuminaMath-CoT~\citep{numina_math_datasets},
DeepScaleR~\citep{deepscaler2025},
OpenThoughts and OpenThoughts3~\citep{guha2025openthoughts},
AIME-style competition problems~\citep{aimo2024aime}, and
OlympiadBench~\citep{he2024olympiadbench}.
After validation, answer normalization, and problem-level deduplication, the
final corpus contains 1,065 unique CR problems, split into 853 training, 106
validation, and 106 test examples.
All trajectories derived from the same underlying problem remain in the same
split to prevent leakage.
Complete prompts and collection settings are provided in
Appendix~\ref{app:prompts}, and the source-level composition is reported in
Appendix~\ref{app:split_statistics}.

\paragraph{Training protocols.}
For supervised fine-tuning, we compare SFT-CR with three matched controls:
SFT-Random, SFT-Already-Solved, and SFT-Wrong.
All variants use the same Qwen3-8B initialization, 616 training examples, LoRA
configuration, and optimization procedure, differing only in the selected
training trajectories.
The 616-example size is determined by the subset of CR training problems with
clean successful blind trajectories.
For on-policy distillation, Qwen3-8B serves as the student and Qwen3-32B as
the frozen teacher.
OPD-CR, OPD-All, and OPD-AlreadySolved use the same student, teacher,
distillation objective, effective batch size, and optimization settings,
differing primarily in the training distribution.
Detailed SFT and OPD configurations are reported in
Appendices~\ref{app:sft_training} and~\ref{app:opd_training}.
\paragraph{Evaluation protocol.}
We use two complementary evaluation settings.
The held-out CR split measures recovery, since every example is a complete
initial failure for the base model.
We additionally evaluate on a 735-example mixed generalization test set to
measure overall performance and regressions on problems the base model already
solves.
This set combines the frozen 106-example CR test split with 629 independently
sampled non-CR problems from the same source pools.
For each source dataset, non-CR examples are sampled according to the natural
proportion of Already-Solved and Unresolved examples in that source pool,
yielding 372 Already-Solved and 257 Unresolved examples.
The construction procedure is detailed in
Appendix~\ref{app:general_test_construction}.
To evaluate compute utilization directly, we test maximum generation budgets
of 512, 1,024, 2,048, and 4,096 tokens under matched decoding settings.
We report accuracy and recovery at each budget under matched deterministic
decoding settings.
All deterministic evaluations are run with batch size 1 to avoid
batching-dependent generation differences observed in our inference stack.

\paragraph{Cross-model evaluation.}
To test whether the CR phenomenon and the benefit of CR-based selection extend
beyond Qwen3-8B, we repeat the core CR construction and matched SFT comparison
on Llama-3.1-8B-Instruct~\citep{dubey2024llama3}.
Each model reconstructs its own CR set from the same underlying reasoning
sources rather than reusing Qwen3-8B labels.
Cross-model CR statistics and additional experimental details are reported in
Appendix~\ref{app:cross_model_details}.

\section{Results}
\label{sec:results}

We first isolate the role of data selection under controlled SFT.
The analysis then moves from held-out CR recovery to an independent test set,
compute-scaling behavior, dataset-level decomposition, OPD, and cross-model
evaluation.

\paragraph{Post-Training Helps, but Data Selection Matters.}
\label{sec:sft_results}
We first ask how much of the recovery gain comes from supervised post-training
itself, and how much depends on which examples are selected for training.
Table~\ref{tab:sft_main} compares four controlled SFT conditions on the
106 held-out CR validation problems.
Base accuracy is 0\% by construction because every example is a complete
low-compute failure for the original model.
\begin{table}[t]
\centering
\caption{
\textbf{Held-out low-budget recovery after controlled SFT.}
All SFT variants use 616 examples and the same base model and LoRA
training configuration.
}
\label{tab:sft_main}

\small
\renewcommand{\arraystretch}{1.08}
\setlength{\tabcolsep}{5pt}

\begin{tabular*}{\linewidth}{
@{\extracolsep{\fill}}
lccccc
}
\toprule

&
\multicolumn{1}{c}{\textit{Base model}}
&
\multicolumn{4}{c}{\textit{Controlled supervised fine-tuning}}
\\

\cmidrule(lr){2-2}
\cmidrule(lr){3-6}

Metric
& Base
& SFT-Wrong
& SFT-Random
& SFT-Already-Solved
& \textbf{SFT-CR}
\\
\midrule

Recovery (\%) $\uparrow$
& 0.0
& 29.2
& 31.1
& 40.6
& \textbf{45.3}
\\

Avg. tokens
& 858
& 888
& 908
& 918
& 891
\\

Trunc. (\%) $\downarrow$
& 0.0
& 10.4
& 10.4
& 13.2
& \textbf{8.5}
\\

\bottomrule
\end{tabular*}

\end{table}
The first result is that post-training itself is already useful.
Even without targeted selection, SFT-Wrong recovers 29.2\% of the original
failures and SFT-Random reaches 31.1\%.
This shows that part of the low-compute failure region can be improved simply
through additional supervised training.
The choice of training data, however, changes the magnitude of that gain.
Training on examples the base model already solves increases recovery to
40.6\%, substantially above random selection.
SFT-CR improves further to 45.3\%, exceeding SFT-Random by 14.2 percentage
points and SFT-Already-Solved by 4.7 points.
Thus, the result is not merely that fine-tuning helps: examples selected by
their response to additional reasoning provide the strongest supervision among
the matched training sets.
The advantage is also not explained by longer generations.
SFT-CR produces 891 tokens on average, fewer than SFT-Random at 908 and
SFT-Already-Solved at 918, while its truncation rate is also lower at 8.5\%.
Checkpoint and multi-seed analyses are reported in
Appendices~\ref{app:sft_training} and~\ref{app:stability}, with additional
efficiency statistics in Appendix~\ref{app:efficiency}.
Post-training therefore recovers a substantial fraction of the original
failures even with generic supervision, but the size of the gain depends
strongly on the selected training data.
We next test whether the advantage of CR selection remains visible outside
the CR-constructed evaluation set.
\paragraph{The Gain Extends beyond the CR-Selected Evaluation Set.}
\label{sec:general_results}
The controlled SFT comparison shows that CR-selected trajectories are more
effective for recovering held-out CR failures than matched alternatives.
However, the held-out CR split is deliberately constructed around the behavior
targeted by CRDS.
We therefore evaluate the same models on a 735-example mixed generalization
test set consisting of the frozen CR test split and independently sampled
Already-Solved and Unresolved examples from the same source pools.
Table~\ref{tab:general_main} reports overall accuracy together with a
stratification according to the base model's responsiveness.
\begin{table}[t]
\centering
\small
\caption{
\textbf{General test accuracy and responsiveness-stratified performance.}
The mixed test set combines the frozen CR test split with independently sampled
Already-Solved and Unresolved examples from the same source pools.
}
\label{tab:general_main}
\setlength{\tabcolsep}{5.5pt}
\renewcommand{\arraystretch}{1.10}

\begin{tabular*}{\linewidth}{
@{\extracolsep{\fill}}
lccc
}
\toprule
Method
& Overall (735)
& CR (106)
& Unresolved (257) \\
\midrule

Base
& 50.6 & 0.0 & 0.0 \\

SFT-Random
& 49.1 & 34.9 & 2.3 \\

SFT-Already-Solved
& 49.4 & 30.2 & 2.7 \\

\textbf{SFT-CR}
& \textbf{51.8}
& \textbf{47.2}
& \textbf{3.9} \\

\bottomrule
\end{tabular*}
\end{table}
As shown in Table~\ref{tab:general_main}, SFT-CR achieves the highest overall
accuracy on the mixed generalization set, improving from 50.6\% for the base
model to 51.8\%.
Because the strata are defined by the base model's initial behavior, the base
row has 100.0\% accuracy on Already-Solved examples and 0.0\% accuracy on both
CR and Unresolved examples by construction.
The overall result is therefore mainly a coarse check that CR training does not
only improve the selected CR split.
The stratified results show where the change occurs.
On compute-responsive problems, SFT-CR increases accuracy from 0.0\% to
47.2\%, compared with 34.9\% for SFT-Random and 30.2\% for
SFT-Already-Solved.
Unresolved problems remain difficult for all methods, with SFT-CR reaching only
3.9\%.
The improvement is therefore concentrated near the recoverable reasoning
boundary rather than distributed uniformly across the test set.
This recovery comes with a retention cost: accuracy on already-solved problems
drops from 100.0\% for the base model to 86.3\% after SFT-CR.
This trade-off is expected because the main comparison uses single-source
matched training sets to isolate the effect of data selection, rather than to
optimize the final deployment mixture.
Appendix~\ref{app:mix_training} shows that a simple 50\% CR + 50\%
Already-Solved mixture raises already-solved accuracy to 87.4\%, above both
SFT-CR at 86.3\% and SFT-Already-Solved at 87.1\%.
The representative cases are
provided in Appendices~\ref{app:case_studies}.
This mixed generalization test shows that CR training is not useful only on the
constructed CR split, while also revealing that most of the gain is
concentrated on recoverable failures.
We next ask whether this improvement also changes how effectively the model
uses different amounts of test-time computation.
\paragraph{CR Training Shifts the Compute--Accuracy Frontier.}
\label{sec:compute_results}
\begin{table}[t]
\centering
\small
\caption{
\textbf{Compute--accuracy scaling across inference budgets.}
Accuracy (\%) is reported under maximum generation budgets from 512 to 4,096 tokens.
}
\label{tab:compute_main}
\setlength{\tabcolsep}{8pt}
\renewcommand{\arraystretch}{1.10}

\begin{tabular*}{0.96\linewidth}{
@{\extracolsep{\fill}}
lcccc
}
\toprule
& \multicolumn{4}{c}{\textbf{Maximum generation budget}} \\
\cmidrule(lr){2-5}
Method & 512 & 1024 & 2048 & 4096 \\
\midrule

Base
& 0.0 & 0.0 & 0.0 & 0.0 \\
SFT-Random
& 9.4 & 24.5 & 34.6 & 37.7 \\
SFT-Already-Solved
& 8.5 & 22.6 & 30.2 & 33.0 \\

\textbf{SFT-CR}
& \textbf{17.0}
& \textbf{28.3}
& \textbf{47.2}
& \textbf{48.1} \\

\bottomrule
\end{tabular*}
\end{table}
The mixed generalization results show that CR training mainly improves
recoverable failures rather than uniformly changing all behavioral regions.
We now test whether this recovery also changes how effectively the model uses
different amounts of test-time computation.
We evaluate all SFT variants at 512, 1,024, 2,048, and 4,096 tokens, with the
controlled comparison reported in Table~\ref{tab:compute_main}.
As shown in Table~\ref{tab:compute_main}, the base model does not recover any
held-out CR examples under the deterministic evaluation protocol, remaining at
0.0\% accuracy across all budgets.
Supervised post-training changes this frontier: both matched controls recover
some failures, but SFT-CR is strongest at every budget.
It reaches 17.0\%, 28.3\%, 47.2\%, and 48.1\% from 512 to 4,096 tokens,
exceeding the strongest non-CR baseline by 7.6, 3.8, 12.6, and 10.4 percentage
points, respectively.
The gain appears both below and above the 2,048-token setting used during CR
construction, indicating that CR training is not merely specialized to the
collection budget.
Instead, CR selection changes the compute--accuracy frontier: the trained model
obtains more accuracy from the same available generation budget than either
random fine-tuning or fine-tuning on already-solved examples.
The 2,048-token evaluation combines examples from several sources, so we next
examine whether this gain is broadly distributed across datasets.
\paragraph{The Frontier Gain Is Supported across Multiple Data Sources.}
\label{sec:dataset_results}
The compute-scaling result at 2,048 tokens aggregates examples from several
reasoning sources.
We therefore decompose this operating point by source dataset to check whether
the SFT-CR gain is driven by a single corpus or supported more broadly.
Table~\ref{tab:dataset_main} reports the corresponding per-source accuracies.
\begin{table*}[t]
\centering
\small
\caption{\textbf{Dataset-wise performance at the 2,048-token evaluation budget.} The final column reports the absolute SFT-CR improvement over the best non-CR SFT baseline in percentage points.}
\resizebox{\textwidth}{!}{%
\begin{tabular}{lrrrrrr}
\toprule
Dataset & $N$ & Base & SFT-Random & SFT-AlreadySolved & SFT-CR & $\Delta$ CR vs.\ Best Non-CR \\
\midrule
NuminaMath-CoT & 47 & 0 & 25.5 & 23.4 & \textbf{40.4} & \textbf{+14.9} \\
DeepScaleR & 27 & 0 & 48.1 & 40.7 & \textbf{51.9} & \textbf{+3.8} \\
OpenThoughts & 14 & 0 & 35.7 & 21.4 & \textbf{57.1} & \textbf{+21.4} \\
OpenThoughts3 & 14 & 0 & 35.7 & 35.7 & \textbf{42.9} & \textbf{+7.2} \\
AIME & 0 & -- & -- & -- & -- & -- \\
OlympiadBench & 4 & 0 & 50.0 & 50.0 & \textbf{75.0} & \textbf{+25.0} \\
\midrule
\textbf{Overall} & \textbf{106} & 0 & 34.9 & 30.2 & \textbf{47.2} & \textbf{+12.3} \\
\bottomrule
\end{tabular}%
}
\label{tab:dataset_main}
\end{table*}
As shown in Table~\ref{tab:dataset_main}, SFT-CR outperforms the matched SFT
controls on every non-empty source subset at the 2,048-token evaluation budget.
The strongest evidence comes from the two largest subsets.
On NuminaMath-CoT, SFT-CR reaches 40.4\%, compared with 25.5\% for
SFT-Random and 23.4\% for SFT-Already-Solved.
On DeepScaleR, SFT-CR reaches 51.9\%, compared with 48.1\% and 40.7\%.
The smaller OpenThoughts and OpenThoughts3 subsets show the same direction,
with SFT-CR improving over the strongest non-CR baseline by 21.4 and 7.2
points, respectively.
OlympiadBench also favors SFT-CR, but it contains only four examples, and no
AIME example appears in this split, so these small-source results should be
treated only as supporting evidence.
Combining all non-empty source subsets, SFT-CR reaches 47.2\% at this budget,
compared with 34.9\% for SFT-Random and 30.2\% for SFT-Already-Solved.
The source-level decomposition therefore supports the aggregate frontier
result: CR training is not benefiting from a single dominant corpus, but
consistently improves recovery across multiple reasoning data sources.
We next test whether CR selection remains useful when the training procedure
itself is no longer based on fixed offline supervision.

\paragraph{CR Selection Also Benefits On-Policy Distillation.}
\label{sec:opd_analysis}
SFT uses a fixed set of training trajectories, so its gains could in principle
come from selecting better static supervision.
We therefore test CR selection under on-policy distillation, where supervision
is instead provided on states visited by the evolving student policy.
We use on-policy distillation (OPD) for this purpose. Unlike ordinary
SFT, where the training targets are fixed before optimization, OPD first samples
trajectories from the current student policy and then uses a stronger teacher
to provide supervision on states that the student actually visits. As training
progresses, the supervised state distribution therefore changes together with
the student. This makes OPD a useful complementary setting for testing CRDS:
if CR selection remains beneficial here, its value cannot be attributed only to
choosing better static SFT targets.
We use Qwen3-8B as the student and Qwen3-32B as the frozen teacher, and compare
OPD-All, OPD-AlreadySolved, and CR-OPD under matched optimization settings.
Table~\ref{tab:opd_main} reports recovery on the held-out CR split for both
single-node and distributed training runs.
\begin{table}[t]
\centering
\small
\caption{
\textbf{Held-out CR recovery under on-policy distillation.}
CR-OPD achieves the highest recovery in both single-node and distributed
training settings.
}
\label{tab:opd_main}
\setlength{\tabcolsep}{5pt}
\renewcommand{\arraystretch}{1.10}

\begin{tabular*}{\linewidth}{
@{\extracolsep{\fill}}
lccc
ccc
}
\toprule
&
\multicolumn{3}{c}{\textbf{Single}}
&
\multicolumn{3}{c}{\textbf{DDP}}
\\
\cmidrule(lr){2-4}
\cmidrule(lr){5-7}

Method
& Rec. $\uparrow$ & Tok. & Trunc.
& Rec. $\uparrow$ & Tok. & Trunc. \\
\midrule

AlreadySolved
& 32.1 & 840 & 9.4
& 30.2 & 884 & 8.5 \\

All
& 33.0 & 947 & 17.9
& 32.1 & 852 & 6.6 \\

\textbf{CR}
& \textbf{37.7} & 871 & 7.5
& \textbf{34.9} & 899 & 13.2 \\

\bottomrule
\end{tabular*}
\end{table}
The same ordering appears in both runs.
Under single-node training, CR-OPD recovers 37.7\% of held-out CR failures,
compared with 33.0\% for OPD-All and 32.1\% for OPD-AlreadySolved.
Under DDP, CR-OPD reaches 34.9\%, again exceeding OPD-All at 32.1\% and
OPD-AlreadySolved at 30.2\%.
The improvement is therefore not specific to one training implementation.
The efficiency statistics also help rule out a simple length-based explanation.
In the single-node run, CR-OPD reaches the highest recovery while generating
871 tokens on average, fewer than OPD-All at 947, and it has a lower truncation
rate of 7.5\% versus 17.9\%.
The DDP run is noisier in truncation behavior, but the recovery ordering remains
unchanged.
This suggests that CR selection improves which student states receive useful
teacher supervision rather than merely encouraging longer generations.
A teacher--student KL diagnostic in Appendix~\ref{app:opd_kl_diagnostic}
supports this interpretation: CR states exhibit the largest initial
teacher--student divergence and the clearest KL decrease during OPD training,
whereas Already-Solved states show low, nearly flat divergence, indicating
limited additional distillation signal.
OPD is therefore a stricter test of the selection principle than ordinary SFT.
Because its supervision already follows the evolving student policy, part of
the training-distribution adaptation is supplied by the objective itself.
The fact that CR-OPD still improves held-out recovery in both runs indicates
that base-model compute responsiveness provides information that is not fully
subsumed by on-policy state selection.
CR-OPD achieves the strongest held-out recovery in both training settings,
showing that the CR signal remains useful even when supervision already adapts
to the student's current policy.
We finally test whether the same pattern extends beyond Qwen3-8B.

\paragraph{Compute Responsiveness Generalizes across Models.}
\label{sec:cross_model}
The preceding experiments establish the effect on Qwen3-8B.
To test whether the same pattern is specific to this model family, we repeat
the CR construction and matched SFT comparison on Llama-3.1-8B-Instruct.
Because CR membership is model specific, each model reconstructs its own CR set
from the same underlying reasoning sources rather than reusing Qwen-derived
labels.
\begin{table}[t]
\centering
\small
\caption{
\textbf{Cross-model generalization.}
CR examples are reconstructed independently for each model, and Random,
Already-Solved, and CR training sets are matched within each model.
}
\label{tab:cross_model}
\setlength{\tabcolsep}{7pt}
\renewcommand{\arraystretch}{1.12}

\begin{tabular*}{\linewidth}{
@{\extracolsep{\fill}}
lccccc
}
\toprule
Model
& Params.
& CR Rate
& Random
& Already
& CR \\
\midrule

Qwen3-8B
& 8B
& 8.3
& 31.1
& 40.6
& \textbf{45.3} \\

Llama-3.1-8B-Instruct
& 8B
& 9.2
& 24.1
& 17.2
& \textbf{27.6} \\

\bottomrule
\end{tabular*}
\end{table}
Table~\ref{tab:cross_model} shows that compute responsiveness is not unique to
Qwen3-8B.
Llama-3.1-8B-Instruct exhibits a comparable CR rate of 9.2\%, slightly higher
than the 8.3\% observed for Qwen3-8B.
More importantly, the same within-model ordering holds for the CR-selected
training data: on Llama-3.1-8B-Instruct, CR-SFT reaches 27.6\% recovery,
compared with 24.1\% for Random and 17.2\% for Already-Solved.
Thus, although the absolute recovery rates differ across models, selecting
examples by compute responsiveness again provides the strongest supervision
among the matched alternatives.
The lower Already-Solved result on Llama-3.1-8B-Instruct also reinforces the
need to reconstruct CR sets separately for each model rather than transferring
labels or assuming that initially solved examples are uniformly useful for
recovering failures.
Appendix~\ref{app:cross_model_details} further shows that CR membership only
partially overlaps across models when reconstructed from the same underlying
reasoning sources.
This fixed-source comparison suggests that CRDS is not selecting a universal
set of easy-to-recover questions, but rather the failures that are recoverable
for a particular model under the specified additional-reasoning protocol.
The cross-model comparison supports the view that CRDS captures a behavioral
training signal that extends beyond a single 8B model family.


\section{Conclusion and Limitations}
\label{sec:conclusion_limitations}

We study test-time compute as a learnable capability rather than only an
allocation problem. By identifying \emph{compute-responsive} examples---problems
that change from complete failure to success under additional reasoning---we
introduce Compute-Responsive Data Selection (CRDS) as a post-training signal
for improving compute utilization. On Qwen3-8B, SFT-CR recovers 45.3\% of
held-out low-compute failures, improves overall accuracy on an independently
sampled test distribution to 51.8\%, and shifts the compute--accuracy frontier
across budgets from 512 to 4,096 tokens. The gains are concentrated near the
model's recoverable reasoning boundary, remain visible under on-policy
distillation, and extend to Llama-3.1-8B-Instruct when CR sets are reconstructed
separately for that model. These results highlight a complementary dimension of
test-time scaling: beyond deciding where to allocate computation, it is also
necessary to improve how effectively the model uses the computation it
receives. The current study is limited to mathematical reasoning, where final
answers can be checked with relatively reliable automatic verification; applying
CRDS to code generation, tool use, scientific reasoning, or open-ended tasks
will require verifiers that can distinguish genuine reasoning recovery from
superficial output changes with comparable reliability. The measured CR set is
also tied to the specific additional-reasoning protocol used here, consisting
of two blind re-solving attempts and two audit-based revisions, so broader
search may reveal additional recoverable examples while increasing collection
cost. Finally, our experiments cover a limited set of post-training objectives
and model families, leaving larger-scale training, stronger reasoning models,
and other optimization methods as important directions for future work.
\subsection*{AI Use Statement}

In preparing this submission, we used large language models (LLMs) as general
writing assistants. Specifically, LLMs were used to improve the clarity and
readability of the manuscript, including sentence restructuring, grammatical
editing, and concision. All research ideas, method design, experiments,
analyses, and final scientific decisions were independently developed,
conducted, and verified by the authors. We reviewed all AI-assisted text and
take responsibility for the final content of this work.

\subsection*{Ethics Statement}

This work does not involve human subjects, private user data, or personally
identifiable information.
The experiments use publicly available mathematical reasoning datasets and
language models to study post-training data selection and reasoning behavior.
The additional reasoning trajectories generated during CR construction are
used solely for research evaluation and model training within this setting.
We follow the applicable licenses and usage conditions of the datasets,
models, and software used in the study.

\subsection*{Reproducibility Statement}

The main text specifies the CR definition, data-selection procedure, training
protocols, evaluation settings, and reported metrics.
The appendix provides the evolution of the CR construction pipeline,
source-level collection and split statistics, complete prompting procedures,
the training-data schema, answer normalization rules, SFT and OPD
configurations, cross-model construction details, multi-seed stability
analysis, CR discovery cost, mixed-training analysis, and construction of the
mixed generalization test set.
Anonymous supplementary materials provide the implementation, evaluation
scripts, and configurations used to reproduce the reported experiments.

\bibliographystyle{iclr2027_conference}
\bibliography{iclr2027_conference}

@inproceedings{wei2022chain,
  title={Chain-of-Thought Prompting Elicits Reasoning in Large Language Models},
  author={Wei, Jason and Wang, Xuezhi and Schuurmans, Dale and Bosma, Maarten and Ichter, Brian and Xia, Fei and Chi, Ed H. and Le, Quoc V. and Zhou, Denny},
  booktitle={Advances in Neural Information Processing Systems},
  year={2022}
}

@inproceedings{wang2023selfconsistency,
  title={Self-Consistency Improves Chain of Thought Reasoning in Language Models},
  author={Wang, Xuezhi and Wei, Jason and Schuurmans, Dale and Le, Quoc V. and Chi, Ed H. and Narang, Sharan and Chowdhery, Aakanksha and Zhou, Denny},
  booktitle={International Conference on Learning Representations},
  year={2023}
}

@inproceedings{lightman2024verify,
  title={Let's Verify Step by Step},
  author={Lightman, Hunter and Kosaraju, Vineet and Burda, Yura and Edwards, Harri and Baker, Bowen and Lee, Teddy and Leike, Jan and Schulman, John and Sutskever, Ilya and Cobbe, Karl},
  booktitle={International Conference on Learning Representations},
  year={2024}
}

@inproceedings{yao2023tree,
  title={Tree of Thoughts: Deliberate Problem Solving with Large Language Models},
  author={Yao, Shunyu and Yu, Dian and Zhao, Jeffrey and Shafran, Izhak and Griffiths, Thomas L. and Cao, Yuan and Narasimhan, Karthik},
  booktitle={Advances in Neural Information Processing Systems},
  year={2023}
}

@inproceedings{madaan2023selfrefine,
  title={Self-Refine: Iterative Refinement with Self-Feedback},
  author={Madaan, Aman and Tandon, Niket and Gupta, Prakhar and Hallinan, Skyler and Gao, Luyu and Wiegreffe, Sarah and Alon, Uri and Dziri, Nouha and Prabhumoye, Shrimai and Yang, Yiming and others},
  booktitle={Advances in Neural Information Processing Systems},
  year={2023}
}

@article{snell2024scaling,
  title={Scaling LLM Test-Time Compute Optimally can be More Effective than Scaling Model Parameters},
  author={Snell, Charlie and Lee, Jaehoon and Xu, Kelvin and Kumar, Aviral},
  journal={arXiv preprint arXiv:2408.03314},
  year={2024}
}

@inproceedings{zhou2023lima,
  title={LIMA: Less Is More for Alignment},
  author={Zhou, Chunting and Liu, Pengfei and Xu, Puxin and Iyer, Srinivasan and Sun, Jiao and Mao, Yuning and Ma, Xuezhe and Efrat, Avia and Yu, Ping and Yu, Lili and Zhang, Susan and Ghosh, Gargi and Lewis, Mike and Zettlemoyer, Luke and Levy, Omer},
  booktitle={Advances in Neural Information Processing Systems},
  year={2023}
}

@inproceedings{wang2023selfinstruct,
  title={Self-Instruct: Aligning Language Models with Self-Generated Instructions},
  author={Wang, Yizhong and Kordi, Yeganeh and Mishra, Swaroop and Liu, Alisa and Smith, Noah A. and Khashabi, Daniel and Hajishirzi, Hannaneh},
  booktitle={Proceedings of the 61st Annual Meeting of the Association for Computational Linguistics},
  year={2023}
}

@article{zhou2023leasttomost,
  title={Least-to-most prompting enables complex reasoning in large language models},
  author={Zhou, Denny and Sch{\"a}rli, Nathanael and Hou, Le and Wei, Jason and Scales, Nathan and Wang, Xuezhi and Schuurmans, Dale and Cui, Claire and Bousquet, Olivier and Le, Quoc and others},
  journal={arXiv preprint arXiv:2205.10625},
  year={2022}
}

@inproceedings{gao2023pal,
  title={PAL: Program-Aided Language Models},
  author={Gao, Luyu and Madaan, Aman and Zhou, Shuyan and Alon, Uri and Liu, Pengfei and Yang, Yiming and Callan, Jamie and Neubig, Graham},
  booktitle={Proceedings of the 40th International Conference on Machine Learning},
  year={2023}
}

@article{chen2023program,
  title={Program of Thoughts Prompting: Disentangling Computation from Reasoning for Numerical Reasoning Tasks},
  author={Chen, Wenhu and Ma, Xueguang and Wang, Xinyi and Cohen, William W.},
  journal={Transactions on Machine Learning Research},
  year={2023}
}

@inproceedings{yao2023react,
  title={ReAct: Synergizing Reasoning and Acting in Language Models},
  author={Yao, Shunyu and Zhao, Jeffrey and Yu, Dian and Du, Nan and Shafran, Izhak and Narasimhan, Karthik and Cao, Yuan},
  booktitle={International Conference on Learning Representations},
  year={2023}
}

@article{shinn2023reflexion,
  title={Reflexion: Language agents with verbal reinforcement learning},
  author={Shinn, Noah and Cassano, Federico and Gopinath, Ashwin and Narasimhan, Karthik and Yao, Shunyu},
  journal={Advances in neural information processing systems},
  volume={36},
  pages={8634--8652},
  year={2023}
}

@inproceedings{liu2020fastbert,
  title={FastBERT: A Self-Distilling BERT with Adaptive Inference Time},
  author={Liu, Weijie and Zhou, Peng and Wang, Zhiruo and Zhao, Zhe and Deng, Haotang and Ju, Qi},
  booktitle={Proceedings of the 58th Annual Meeting of the Association for Computational Linguistics},
  year={2020}
}

@inproceedings{schuster2022calm,
  title={Confident Adaptive Language Modeling},
  author={Schuster, Tal and Fisch, Adam and Gupta, Jai and Dehghani, Mostafa and Bahri, Dara and Tran, Vinh Q. and Tay, Yi and Metzler, Donald},
  booktitle={Advances in Neural Information Processing Systems},
  year={2022}
}

@inproceedings{schwartz2020righttool,
  title={The Right Tool for the Job: Matching Model and Instance Complexities},
  author={Schwartz, Roy and Stanovsky, Gabriel and Swayamdipta, Swabha and Dodge, Jesse and Smith, Noah A.},
  booktitle={Proceedings of the 58th Annual Meeting of the Association for Computational Linguistics},
  year={2020}
}

@article{chen2024frugalgpt,
  title={FrugalGPT: How to Use Large Language Models While Reducing Cost and Improving Performance},
  author={Chen, Lingjiao and Zaharia, Matei and Zou, James},
  journal={Transactions on Machine Learning Research},
  year={2024}
}

@inproceedings{ong2025routellm,
  title={RouteLLM: Learning to Route LLMs from Preference Data},
  author={Ong, Isaac and Almahairi, Amjad and Wu, Vincent and Chiang, Wei-Lin and Wu, Tianhao and Gonzalez, Joseph E. and Kadous, M. Waleed and Stoica, Ion},
  booktitle={International Conference on Learning Representations},
  year={2025}
}

@inproceedings{rafailov2023dpo,
  title={Direct Preference Optimization: Your Language Model is Secretly a Reward Model},
  author={Rafailov, Rafael and Sharma, Archit and Mitchell, Eric and Ermon, Stefano and Manning, Christopher D. and Finn, Chelsea},
  booktitle={Advances in Neural Information Processing Systems},
  volume={36},
  year={2023}
}

@article{shao2024deepseekmath,
  title={DeepSeekMath: Pushing the Limits of Mathematical Reasoning in Open Language Models},
  author={Shao, Zhihong and Wang, Peiyi and Zhu, Qihao and Xu, Runxin and Song, Junxiao and Bi, Xiao and Zhang, Haowei and Zhang, Mingchuan and Li, Y. K. and Wu, Y. and Guo, Daya},
  journal={arXiv preprint arXiv:2402.03300},
  year={2024}
}

@inproceedings{hsieh2023distilling,
  title={Distilling Step-by-Step! Outperforming Larger Language Models with Less Training Data and Smaller Model Sizes},
  author={Hsieh, Cheng-Yu and Li, Chun-Liang and Yeh, Chih-Kuan and Nakhost, Hootan and Fujii, Yasuhisa and Ratner, Alexander and Krishna, Ranjay and Lee, Chen-Yu and Pfister, Tomas},
  booktitle={Findings of the Association for Computational Linguistics: ACL 2023},
  pages={8003--8017},
  year={2023}
}

@inproceedings{agarwal2024onpolicy,
  title={On-Policy Distillation of Language Models: Learning from Self-Generated Mistakes},
  author={Agarwal, Rishabh and Vieillard, Nino and Zhou, Yongchao and Stanczyk, Piotr and Garea, Sabela Ramos and Geist, Matthieu and Bachem, Olivier},
  booktitle={International Conference on Learning Representations},
  year={2024}
}

@article{bai2022constitutional,
  title={Constitutional ai: Harmlessness from ai feedback},
  author={Bai, Yuntao and Kadavath, Saurav and Kundu, Sandipan and Askell, Amanda and Kernion, Jackson and Jones, Andy and Chen, Anna and Goldie, Anna and Mirhoseini, Azalia and McKinnon, Cameron and others},
  journal={arXiv preprint arXiv:2212.08073},
  year={2022}
}

@inproceedings{welleck2023selfcorrect,
  title={Generating Sequences by Learning to Self-Correct},
  author={Welleck, Sean and Lu, Ximing and West, Peter and Brahman, Faeze and Shen, Tianxiao and Khashabi, Daniel and Choi, Yejin},
  booktitle={International Conference on Learning Representations},
  year={2023}
}

@inproceedings{gou2024critic,
  title={{CRITIC}: Large Language Models Can Self-Correct with Tool-Interactive Critiquing},
  author={Gou, Zhibin and Shao, Zhihong and Gong, Yeyun and Shen, Yelong and Yang, Yujiu and Duan, Nan and Chen, Weizhu},
  booktitle={International Conference on Learning Representations},
  year={2024}
}

@misc{qwen3technicalreport,
  title={Qwen3 Technical Report},
  author={{Qwen Team}},
  year={2025},
  eprint={2505.09388},
  archivePrefix={arXiv},
  primaryClass={cs.CL},
}

@misc{numina_math_datasets,
  title={Numinamath: The largest public dataset in ai4maths with 860k pairs of competition math problems and solutions},
  author={Li, Jia and Beeching, Edward and Tunstall, Lewis and Lipkin, Ben and Soletskyi, Roman and Huang, Shengyi and Rasul, Kashif and Yu, Longhui and Jiang, Albert Q and Shen, Ziju and others},
  journal={Hugging Face repository},
  volume={13},
  number={9},
  pages={9},
  year={2024}
}

@misc{deepscaler2025,
  title={Deepscaler: Surpassing o1-preview with a 1.5 b model by scaling rl},
  author={Luo, Michael and Tan, Sijun and Wong, Justin and Shi, Xiaoxiang and Tang, William Y and Roongta, Manan and Cai, Colin and Luo, Jeffrey and Li, Li Erran and Popa, Raluca Ada and others},
  journal={Notion Blog},
  volume={3},
  number={5},
  year={2025}
}

@misc{guha2025openthoughts,
  title={OpenThoughts: Data Recipes for Reasoning Models},
  author={Guha, Etash and Marten, Ryan and Keh, Sedrick and Raoof, Negin and Smyrnis, Georgios and Bansal, Hritik and Nezhurina, Marianna and Mercat, Jean and Vu, Trung and Sprague, Zayne and Suvarna, Ashima and Feuer, Benjamin and Chen, Liangyu and Khan, Zaid and Frankel, Eric and Grover, Sachin and Choi, Caroline and Muennighoff, Niklas and Su, Shiye and Zhao, Wanjia and others},
  year={2025},
  eprint={2506.04178},
  archivePrefix={arXiv},
  primaryClass={cs.CL},
}

@inproceedings{he2024olympiadbench,
  title={OlympiadBench: A Challenging Benchmark for Promoting AGI with Olympiad-Level Bilingual Multimodal Scientific Problems},
  author={He, Chaoqun and Luo, Renjie and Bai, Yuzhuo and Hu, Shengding and Thai, Zhen Leng and Shen, Junhao and Hu, Jinyi and Han, Xu and Huang, Yujie and Zhang, Yuxiang and Liu, Jie and Qi, Lei and Liu, Zhiyuan and Sun, Maosong},
  booktitle={Proceedings of the 62nd Annual Meeting of the Association for Computational Linguistics},
  year={2024}
}

@article{dubey2024llama3,
  title={The Llama 3 Herd of Models},
  author={Dubey, Abhimanyu and Jauhri, Abhinav and Pandey, Abhinav and Kadian, Abhishek and Al-Dahle, Ahmad and Letman, Aiesha and Mathur, Akhil and Schelten, Alan and Yang, Amy and Fan, Angela and Goyal, Anirudh and Hartshorn, Anthony and Yang, Aobo and Mitra, Archi and Sravankumar, Archie and Korenev, Artem and Hinsvark, Arthur and Rao, Arun and Zhang, Aston and Rodriguez, Aurelien and Gregerson, Austen and Spataru, Ava and Roziere, Baptiste and Biron, Bethany and Tang, Binh and Chern, Bobbie and Caucheteux, Charlotte and Nayak, Chaya and Bi, Chloe and Marra, Chris and McConnell, Chris and Keller, Christian and Touret, Christophe and Wu, Chunyang and Wong, Corinne and Ferrer, Cristian Canton and Nikolaidis, Cyrus and others},
  journal={arXiv preprint arXiv:2407.21783},
  year={2024}
}

@misc{aimo2024aime,
  author={{AI-MO}},
  title={{AIMO Validation AIME}},
  year={2024},
  howpublished={Hugging Face dataset},
  url={https://huggingface.co/datasets/AI-MO/aimo-validation-aime},
  note={Accessed: 2026-09-25}
}

@article{wen2026not,
  title={Not All Errors Are Equal: Consequence-Aware Reasoning Compute Allocation},
  author={Wen, Jingbo and He, Liang and He, Ziqi},
  journal={arXiv preprint arXiv:2606.04402},
  year={2026}
}
\appendix

\clearpage

\noindent{\large\scshape Appendix Contents}

\vspace{1.2em}

\newcommand{\appcontentsline}[3]{%
  \noindent
  \makebox[2em][l]{#1}%
  #2\ \dotfill\ \pageref{#3}\par
  \vspace{0.35em}
}

\appcontentsline{A}
{Evolution of the Compute-Responsive Data-Collection Pipeline}
{app:pipeline_evolution}

\appcontentsline{B}
{Data Collection Statistics}
{app:collection_statistics}

\appcontentsline{C}
{Frozen Train/Validation/Test Composition}
{app:split_statistics}

\appcontentsline{D}
{Prompting and Additional-Reasoning Procedures}
{app:prompts}

\appcontentsline{E}
{Training Data Schema}
{app:data_schema}

\appcontentsline{F}
{Answer Normalization and Format Artifacts}
{app:format_artifacts}

\appcontentsline{G}
{SFT Training Details}
{app:sft_training}

\appcontentsline{H}
{OPD Training Details}
{app:opd_training}

\appcontentsline{I}
{Inference and Training Efficiency}
{app:efficiency}

\appcontentsline{J}
{Qualitative Recovery and Regression Cases}
{app:case_studies}

\appcontentsline{K}
{Cross-Model Collection and Overlap}
{app:cross_model_details}

\appcontentsline{L}
{Training Stability}
{app:stability}

\appcontentsline{M}
{First-Hit Cost of CR Discovery}
{app:first_hit_cost}

\appcontentsline{N}
{Retention--Recovery Trade-off under Mixed Training}
{app:mix_training}

\appcontentsline{O}
{Teacher--Student KL Diagnostic in OPD}
{app:opd_kl_diagnostic}

\appcontentsline{P}
{Construction of the Mixed Generalization Test Set}
{app:general_test_construction}

\clearpage

\section{Evolution of the Compute-Responsive Data-Collection Pipeline}
\label{app:pipeline_evolution}

The final CR construction procedure was not assumed at the beginning of the
project. It emerged from a sequence of pilot experiments that progressively
separated completion effects, reasoning revision, sampling diversity, and
evaluation artifacts. We retain this development because it clarifies why the
final definition of compute responsiveness is more restrictive than a simple
low-wrong/high-correct rule.

\paragraph{Initial Token-Cap Experiments}

Our initial implementation defined low and high compute using the same problem,
prompt, model, and greedy decoding configuration while increasing
\texttt{max\_new\_tokens}. Conceptually, the intended transition was
\begin{equation}
\text{low-budget failure}
\rightarrow
\text{high-budget success}.
\end{equation}
Early experiments showed that this definition frequently selected examples in
which the low output reached the generation cap before completing the
solution. Increasing the cap allowed the same trajectory to continue and
sometimes reach the correct answer.
This behavior is meaningful for understanding generation-length requirements,
but it does not demonstrate a change in reasoning quality. We therefore refer
to it as a \emph{completion boundary}. The pilot also revealed the converse:
when greedy decoding terminates naturally before the low budget is exhausted,
increasing the maximum generation length does not force the model to continue
reasoning. Once EOS has been produced, a larger cap alone does not generate a
new trajectory.

This observation motivated the distinction between
$D_{\mathrm{CR}}^{\mathrm{completion}}$ and
$D_{\mathrm{CR}}^{\mathrm{reasoning}}$. Completion-responsive examples satisfy
\begin{equation}
s^{\mathrm{low}}=0,
\quad
t^{\mathrm{low}}=1,
\quad
s^{\mathrm{high}}=1,
\quad
t^{\mathrm{high}}=0,
\end{equation}
whereas reasoning-responsive examples require the initial output itself to be
complete:
\begin{equation}
s^{\mathrm{low}}=0,
\quad
t^{\mathrm{low}}=0,
\quad
s^{\mathrm{high}}=1.
\end{equation}
Only the latter are used for the primary CR reasoning experiments.

\paragraph{From More Tokens to Explicit Additional Reasoning}

Once truncated failures were removed, simply changing the token cap produced
few genuine low-complete-wrong to high-correct transitions. We therefore
redefined high compute as an additional reasoning procedure rather than
additional unused token capacity.
The first version used audit-first revision. The model received the problem
and previous solution and was instructed to search for mistakes, diagnose the
first unreliable step, re-solve the problem, compare the two solutions, and
revise the answer if necessary. This transformed high compute from
\begin{equation}
\text{same prompt}+\text{larger token cap}
\end{equation}
into
\begin{equation}
\text{initial solve}
+
\text{error search}
+
\text{re-solving}
+
\text{revision}.
\end{equation}

The high maximum generation length of 4,096 tokens was retained only to ensure
that the revision procedure had enough space to finish. The conceptual source
of additional computation was the second reasoning process, not the larger cap
itself.

\paragraph{Why Deterministic Audit Was Insufficient}

A second pilot revealed that deterministic audit frequently reproduced the
same reasoning path as the original solution. Under greedy decoding, the model
would inspect an incorrect derivation, fail to identify the crucial mistake,
then re-solve the problem using essentially the same intermediate steps and
confirm the same wrong answer.

Inspection of failed examples showed several distinct behaviors. In some
cases, the audit reported that no error could be found and reproduced the
original incorrect reasoning. In other cases, the model correctly identified
an unreliable assumption but still failed during the subsequent re-solve.
These observations suggested that a single deterministic revision path was a
poor probe of whether the model possessed any recoverable solution path.
We therefore changed the high-compute stage from one deterministic revision to
multiple sampled reasoning attempts. The low pass remained greedy so that the
initial failure state was stable and reproducible. High reasoning used
sampling with temperature $0.6$--$0.7$ and top-$p=0.9$, allowing alternative
reasoning paths to be explored.

\paragraph{Combining Blind Re-Solving and Audit-Based Revision}

Sampling alone does not remove anchoring to the previous solution when the
model is explicitly shown the failed trajectory. We therefore divided the
high-compute attempts into two complementary modes.
Blind re-solving receives only the original problem. It asks whether the model
can recover through an independent reasoning trajectory without being anchored
to the previous solution. Audit-based revision receives both the problem and
the previous trajectory, allowing the model to inspect and explicitly correct
the first attempt.
The final Qwen3-8B collection procedure uses two blind attempts and two audit
attempts. A candidate is considered recoverable if any of these additional
reasoning paths produces a validated correct answer without truncation.
This changes the interpretation of the high-compute stage from a single
successful generation to the existence of a recoverable path:
\begin{equation}
\mathrm{Recoverable}(x)
=
\mathbb{I}
\left[
\exists y \in \mathcal{H}(x):
\mathrm{Eval}(y)=1
\right].
\end{equation}

\paragraph{The Grader-Format Artifact}

An important third refinement arose from manual inspection of apparently
successful CR examples. One early example involved a concurrent-lines problem
whose mathematically correct answer was $-2/7$. The low output returned
``$k=-2/7$'', whereas one high output returned ``$-2/7$''. A brittle grader
classified the former as incorrect and the latter as correct even though the
mathematical conclusion was unchanged.
Similar artifacts appeared in coordinate expressions, multiple-choice labels,
units, and LaTeX spacing. For example,
\begin{equation}
\left(0,\frac{1}{8} \right)
\quad\text{and}\quad
\left(0,\frac{1}{8}\right)
\end{equation}
could be treated differently by string-based matching despite being
mathematically identical.

This analysis showed that a raw change in grader outcome was insufficient for
assigning reasoning responsiveness. We therefore strengthened answer
normalization to remove variable-name prefixes, option-label wrappers, units,
irrelevant natural-language packaging, LaTeX spacing, and other superficial
differences before comparing mathematical content.
Examples in which low and high outputs are mathematically equivalent after
normalization are treated as formatting corrections rather than reasoning
changes. Complex expressions for which semantic equivalence is uncertain are
retained separately for manual or stronger-verifier inspection.

\paragraph{Final Collection Criterion}

The final reasoning-responsive corpus therefore requires a complete initial
failure, a validated successful additional reasoning trajectory, and
semantic answer normalization that rules out purely formatting-induced
success. In practice, the final filter is
\begin{equation}
s_i^{\mathrm{low}}=0,
\quad
t_i^{\mathrm{low}}=0,
\quad
\exists y \in \mathcal{H}_i:
\mathrm{Eval}(y)=1,
\quad
t(y)=0.
\end{equation}

The evolution from token-cap scaling to sampled multi-path recovery was not
merely an engineering refinement. It changes the scientific interpretation of
the selected data. The final corpus does not represent examples for which the
model simply needed more space to finish writing; it represents examples for
which an additional reasoning process can recover a complete failed attempt.

\paragraph{Completion-Responsive Examples}

Completion-responsive examples are retained separately because they represent
a distinct form of test-time compute dependence. These examples satisfy
\begin{equation}
s_i^{\mathrm{low}}=0,
\qquad
t_i^{\mathrm{low}}=1,
\qquad
s_i^{\mathrm{high}}=1,
\qquad
t_i^{\mathrm{high}}=0.
\end{equation}

Such examples indicate that the low budget is insufficient to complete the
reasoning trajectory. They may be useful for studying reasoning compression or
generation-length adaptation, but they do not isolate reasoning revision.
The main experiments therefore exclude them from $D_{\mathrm{CR}}$.


\section{Data Collection Statistics}
\label{app:collection_statistics}

Table~\ref{tab:collection_full} reports the intermediate large-scale collection
statistics that motivated the choice of data sources. These counts precede the
final continuation, normalization, and problem-level deduplication that produce
the frozen 1,065-example V1 corpus.

\begin{table}[t]
\centering
\small
\caption{
Compute-responsive examples observed during the large-scale collection stage.
}
\label{tab:collection_full}
\setlength{\tabcolsep}{11pt}
\renewcommand{\arraystretch}{1.10}

\begin{tabular*}{0.88\linewidth}{
@{\extracolsep{\fill}}
lrrr
}
\toprule
Dataset & Evaluated & CR & CR Rate \\
\midrule

AIME             & 90    & 4   & 4.4\% \\
OlympiadBench    & 674   & 35  & 5.2\% \\
OpenThoughts     & 2,000 & 145 & 7.2\% \\
OpenThoughts3    & 2,000 & 142 & 7.1\% \\
DeepScaleR       & 3,000 & 269 & 9.0\% \\
NuminaMath-CoT   & 5,067 & 470 & 9.3\% \\

\midrule

\textbf{Observed total}
& \textbf{12,831}
& \textbf{1,065}
& \textbf{8.3\%} \\

\bottomrule
\end{tabular*}
\end{table}
The substantial variation in Table~\ref{tab:collection_full} further supports
the distinction between compute responsiveness and difficulty. AIME and
OlympiadBench are challenging datasets, yet they exhibit relatively low CR
rates because many initial failures remain unresolved. NuminaMath-CoT contains
a larger fraction of problems near the recoverable boundary and consequently
produces substantially more CR examples.

\section{Frozen Train/Validation/Test Composition}
\label{app:split_statistics}

\begin{figure}[t]
\centering
\includegraphics[width=\linewidth]{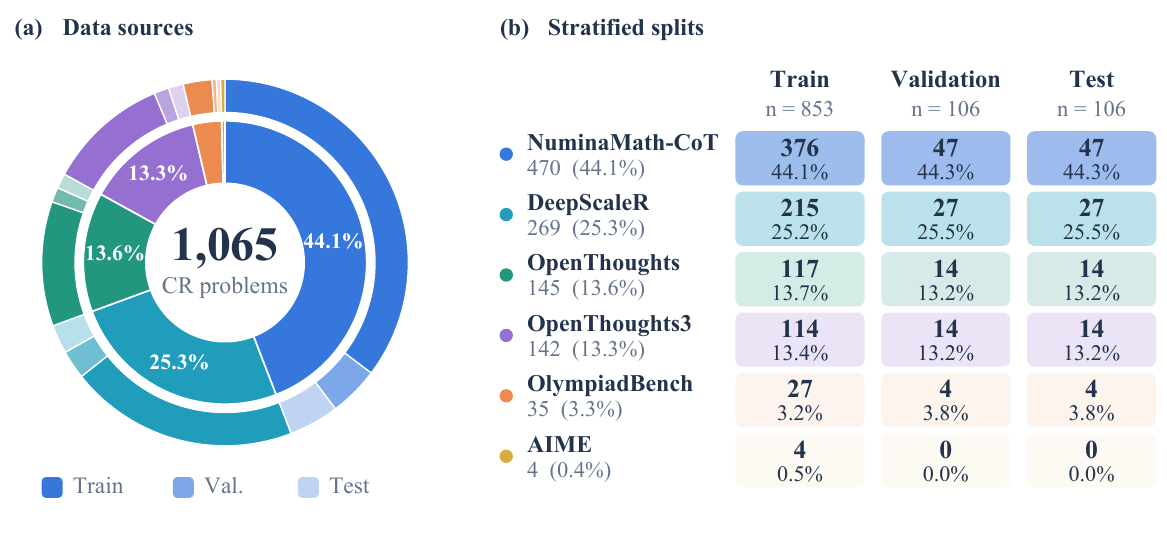}
\caption{
\textbf{Composition of the frozen CR corpus.}
(a) Source composition of the 1,065 compute-responsive problems and their
train, validation, and test allocation.
(b) Stratified source composition across the frozen training, validation, and
test splits.
}
\label{fig:appendix_composition}
\end{figure}

After final validation and problem-level deduplication, we freeze 1,065 unique
CR problems and split them at the problem level into 853 training, 106
validation, and 106 test examples. As shown in
Table~\ref{tab:split_full}, the split is stratified by source dataset using a
fixed seed, preserving nearly identical source proportions across the three
subsets; Figure~\ref{fig:appendix_composition} visualizes the corresponding
source and split composition.





\begin{table}[t]
\centering
\small
\caption{
Per-dataset composition of the frozen V1 CR corpus.
Percentages are computed within each split.
}
\label{tab:split_full}
\setlength{\tabcolsep}{4.5pt}
\renewcommand{\arraystretch}{1.08}

\begin{tabular*}{\linewidth}{
@{\extracolsep{\fill}}
l
rr
rr
rr
rr
}
\toprule
&
\multicolumn{2}{c}{\textbf{Train}}
&
\multicolumn{2}{c}{\textbf{Validation}}
&
\multicolumn{2}{c}{\textbf{Test}}
&
\multicolumn{2}{c}{\textbf{Total}}
\\
\cmidrule(lr){2-3}
\cmidrule(lr){4-5}
\cmidrule(lr){6-7}
\cmidrule(lr){8-9}

Dataset
& \(N\) & \%
& \(N\) & \%
& \(N\) & \%
& \(N\) & \% \\
\midrule

NuminaMath-CoT
& 376 & 44.1
& 47 & 44.3
& 47 & 44.3
& 470 & 44.1 \\

DeepScaleR
& 215 & 25.2
& 27 & 25.5
& 27 & 25.5
& 269 & 25.3 \\

OpenThoughts
& 117 & 13.7
& 14 & 13.2
& 14 & 13.2
& 145 & 13.6 \\

OpenThoughts3
& 114 & 13.4
& 14 & 13.2
& 14 & 13.2
& 142 & 13.3 \\

AIME
& 4 & 0.5
& 0 & 0.0
& 0 & 0.0
& 4 & 0.4 \\

OlympiadBench
& 27 & 3.2
& 4 & 3.8
& 4 & 3.8
& 35 & 3.3 \\

\midrule

\textbf{Total}
& \textbf{853} & \textbf{100.0}
& \textbf{106} & \textbf{100.0}
& \textbf{106} & \textbf{100.0}
& \textbf{1,065} & \textbf{100.0} \\

\bottomrule
\end{tabular*}
\end{table}
Table~\ref{tab:split_full} shows that the source distribution is highly
consistent across the three splits.
NuminaMath-CoT forms the largest portion of the corpus, accounting for 44.1\%
of the full CR set, followed by DeepScaleR at 25.3\%, OpenThoughts at 13.6\%,
and OpenThoughts3 at 13.3\%.
OlympiadBench contributes 3.3\%, while only four AIME problems remain in the
final CR corpus, so no AIME examples appear in the validation or test split.
All trajectories derived from the same underlying problem are grouped before
splitting. Thus, low-compute, blind re-solving, audit, and repair trajectories
from one problem always remain in the same subset, preventing trajectory-level
leakage between training, validation, and test.
\section{Prompting and Additional-Reasoning Procedures}
\label{app:prompts}

The low-compute pass receives only the original problem under the standard
Qwen chat template. It uses greedy decoding with
\texttt{max\_new\_tokens=2048} and
\texttt{enable\_thinking=False}.
Blind high-compute attempts receive the same problem without the previous
solution and use sampling so that independent reasoning paths can be explored.
Audit attempts receive the problem and previous reasoning in the same user
message. The final audit template has the following conceptual structure.

\begin{quote}
\small
\textbf{Problem:} [PROBLEM]

\textbf{A previous solution attempt is shown below:}

[LOW SOLUTION]

\textbf{Stage 1: Error search.}
Carefully inspect the previous solution and determine whether any condition,
algebraic transformation, arithmetic computation, assumption, case analysis,
or final-answer interpretation may be incorrect.

\textbf{Stage 2: Error diagnosis.}
If a mistake is found, identify the earliest unreliable step and explain why it
is problematic. If no mistake is found, still verify the solution
independently.

\textbf{Stage 3: Re-solve.}
Solve the problem again, preferably using an independent reasoning path and
avoiding any suspected error.

\textbf{Stage 4: Compare and revise.}
Compare the new solution with the previous one and output the answer best
supported by the problem conditions and derivation.
\end{quote}

The prompt does not state that the previous solution is wrong. Although
candidate problems have already been judged incorrect by the external
verifier, telling the model that the previous solution must be wrong can induce
unnecessary answer changes. The audit is therefore framed as an independent
verification problem.
The final high-compute procedure uses sampled blind and audit attempts with
temperature approximately $0.6$--$0.7$, top-$p=0.9$, and a maximum generation
budget sufficient to complete the revision procedure. The main Qwen
collection uses two blind and two audit attempts per candidate.

\section{Training Data Schema}
\label{app:data_schema}

Each retained CR example stores enough information to reproduce both the
initial failure and the successful additional-reasoning trajectory. The
important principle is that \texttt{low\_output} and
\texttt{repair\_output} contain the complete generated text rather than only
the extracted final answer.
\begin{verbatim}
{
  "id": "...",
  "dataset": "...",
  "split": "train",

  "problem": "...",
  "gold_answer": "...",
  "gold_answer_normalized": "...",

  "low_prompt": "...",
  "low_output": "...",
  "low_answer": "...",
  "low_answer_normalized": "...",
  "low_correct": false,
  "low_truncated": false,
  "low_tokens": 1735,

  "repair_prompt": "...",
  "repair_output": "...",
  "repair_answer": "...",
  "repair_answer_normalized": "...",
  "repair_correct": true,
  "repair_truncated": false,
  "repair_tokens": 2468,

  "repair_attempt_id": 1,
  "num_repair_attempts": 4,
  "repair_decoding": "sampling",
  "repair_temperature": 0.7,
  "repair_top_p": 0.9,

  "repair_model": "Qwen3-8B",
  "repair_success_source": "blind",
  "enable_thinking": false,
  "is_training_eligible": true
}
\end{verbatim}
When multiple additional attempts succeed, the training dataset contains only
one selected target trajectory per problem. This prevents problems with more
successful attempts from being implicitly up-weighted.

\section{Answer Normalization and Format Artifacts}
\label{app:format_artifacts}

The original grader produced several false low-to-high transitions caused by
answer formatting rather than mathematical changes. A representative example
involved the mathematically identical answers
\begin{equation}
k=-\frac{2}{7}
\qquad\text{and}\qquad
-\frac{2}{7}.
\end{equation}
The first was initially rejected because it contained a variable prefix while
the second matched the gold answer directly. Similar artifacts arose from
units, multiple-choice labels, natural-language wrappers, and LaTeX formatting,
as illustrated in Table~\ref{tab:normalization_examples}.
We therefore normalize predictions before assigning CR labels. Normalization
removes semantically irrelevant wrappers when possible and compares
mathematical content rather than raw strings. Format-only transitions are
stored separately and do not serve as evidence of reasoning recovery.

\begin{table}[t]
\centering
\small
\caption{
Examples of normalization rules introduced after the grader audit.
}
\label{tab:normalization_examples}
\setlength{\tabcolsep}{12pt}
\renewcommand{\arraystretch}{1.15}

\begin{tabular*}{0.92\linewidth}{
@{\extracolsep{\fill}}
lll
}
\toprule
Raw form A & Raw form B & Normalized meaning \\
\midrule

$k=-2/7$
& $-2/7$
& $-2/7$ \\

$\left(0,\frac18 \right)$
& $\left(0,\frac18\right)$
& $\left(0,\frac18\right)$ \\

``C: $1/2$''
& $1/2$
& $1/2$ \\

``80 cm$^2$''
& $80$
& $80$ \\

\bottomrule
\end{tabular*}
\end{table}

\section{SFT Training Details}
\label{app:sft_training}

The controlled SFT experiments use Qwen3-8B with LoRA. The primary
configuration uses rank $r=16$ and scaling parameter $\alpha=32$, with loss
computed only on completion tokens. The optimizer uses a learning rate of
$2\times 10^{-4}$ with cosine decay and an effective batch size of eight.
All SFT variants are trained using the same optimization protocol, and
checkpoints are selected using the frozen validation split.
Table~\ref{tab:sft_data} summarizes the matched training sets used in the
controlled comparison.
\begin{table}[t]
\centering
\small
\caption{
Controlled SFT training sets.
}
\label{tab:sft_data}
\setlength{\tabcolsep}{10pt}
\renewcommand{\arraystretch}{1.12}

\begin{tabular*}{0.94\linewidth}{
@{\extracolsep{\fill}}
lrl
}
\toprule
Method & Examples & Target \\
\midrule

SFT-Wrong
& 616
& Incorrect low trajectory \\

SFT-Random
& 616
& Random correct trajectory \\

SFT-Already-Solved
& 616
& Correct low trajectory \\

\textbf{SFT-CR}
& \textbf{616}
& \textbf{Successful CR blind trajectory} \\

\bottomrule
\end{tabular*}
\end{table}
The final Qwen CR-SFT run stores checkpoints after each epoch.
Table~\ref{tab:sft_checkpoints} reports the available validation results.
Epoch 2 achieves the best observed recovery rate of 45.3\%, while Epoch 3
remains close at 42.5\%, suggesting that the gain is not tied to a single
isolated checkpoint.
\begin{table}[t]
\centering
\small
\caption{
CR-SFT checkpoint selection on the frozen validation set.
}
\label{tab:sft_checkpoints}
\setlength{\tabcolsep}{11pt}
\renewcommand{\arraystretch}{1.12}

\begin{tabular*}{0.88\linewidth}{
@{\extracolsep{\fill}}
lrrr
}
\toprule
Checkpoint & Recovery & Avg. Tokens & Truncation \\
\midrule

Epoch 1
& 41.5\% & 865 & 7.5\% \\

\textbf{Epoch 2}
& \textbf{45.3\%}
& \textbf{891}
& \textbf{8.5\%} \\

Epoch 3
& 42.5\% & 894 & 8.5\% \\

\bottomrule
\end{tabular*}
\end{table}

\section{OPD Training Details}
\label{app:opd_training}

The on-policy distillation (OPD) experiments use Qwen3-8B as the student and
Qwen3-32B as the teacher. The student is trained with LoRA using rank $r=16$,
scaling parameter $\alpha=32$, and dropout $0.05$, applied to all attention and
MLP projections ($q,k,v,o$ and gate/up/down); the teacher is kept frozen and
loaded in 4-bit NF4 with bf16 compute. Because both models share the same
$151{,}936$-token vocabulary, the token-level divergence between them is well
defined.

Each training step is on-policy. Given a prompt, the student first generates a
rollout by sampling with temperature $0.9$ and nucleus $p=0.9$ for up to
$1{,}024$ new tokens. The student and the frozen teacher then score this exact
sequence, and the student is updated with a forward-KL distillation loss
evaluated only on the generated tokens,
\begin{equation}
\mathcal{L}_{\mathrm{OPD}}
= \frac{1}{|G|}\sum_{t\in G}
\mathrm{KL}\!\big(p^{\mathrm{teacher}}(\cdot\mid x_{<t})
\,\big\|\,p^{\mathrm{student}}_{\theta}(\cdot\mid x_{<t})\big),
\end{equation}
with distillation temperature $T=1$ and $G$ the set of generated token
positions. Gradients flow only through the student LoRA parameters; the prompt
tokens and the teacher are excluded from the loss. Consequently the supervision
signal is always defined on states that the current student policy actually
visits, which is the property OPD is designed to test.

Optimization uses AdamW with a learning rate of $1\times10^{-4}$, a cosine
schedule with $3\%$ warmup, and gradient clipping at $1.0$. The effective batch
size is fixed at eight and each condition is trained for a single epoch, giving
$77$ optimizer steps over the $616$ matched prompts. OPD-All and CR-OPD share an
identical training and optimization protocol and differ only in the prompt set
supplied to the rollout stage. We report two implementations of this
configuration: a single-node run with gradient accumulation of eight, and an
eight-node distributed run with gradient accumulation of one that preserves the
same effective batch. The two produce matching recovery within noise, with the
distributed run roughly six times faster. Evaluation follows the same protocol
as the SFT conditions: the final student adapter is evaluated on the frozen
validation split under greedy decoding with a $2{,}048$-token budget.

\section{Inference and Training Efficiency}
\label{app:efficiency}

In addition to accuracy and recovery, we record generated tokens, truncation,
wall-clock inference time, and training time.
Table~\ref{tab:efficiency_full} summarizes the corresponding efficiency
statistics for the SFT and OPD variants.
\begin{table*}[t]
\centering
\small
\caption{
Training and low-budget inference efficiency.
Missing latency values will be filled from the unified timing evaluation.
}
\label{tab:efficiency_full}
\setlength{\tabcolsep}{5pt}
\renewcommand{\arraystretch}{1.10}

\resizebox{0.98\textwidth}{!}{%
\begin{tabular}{lcccccc}
\toprule
Method
& Recovery
& Avg. Tokens
& \shortstack{Latency /\\Sample}
& \shortstack{Total Eval.\\Time}
& Truncation
& Training Time \\
\midrule

\rowcolor{gray!10}
\multicolumn{7}{l}{\textit{Base model}} \\
Base
& 0.0 & 858 & -- & -- & 0.0 & -- \\

\addlinespace[2pt]
\rowcolor{gray!10}
\multicolumn{7}{l}{\textit{Supervised fine-tuning}} \\
SFT-Wrong
& 29.2 & 888 & 5.6 s & 9.9 min & 10.4 & $\sim$20 min \\
SFT-Random
& 31.1 & 908 & 5.5 s & 9.8 min & 10.4 & $\sim$20 min \\
SFT-Already-Solved
& 40.6 & 918 & 5.5 s & 9.6 min & 13.2 & $\sim$20 min \\
\textbf{SFT-CR}
& \textbf{45.3} & 891 & 5.8 s & 10.2 min & \textbf{8.5} & $\sim$20 min \\

\addlinespace[2pt]
\rowcolor{gray!10}
\multicolumn{7}{l}{\textit{On-policy distillation}} \\
OPD-All
& 33.0 & 947 & 5.8 s & 10.3 min & 17.9 & $\sim$10.5 h \\
OPD-Already-Solved
& 32.1 & 840 & 6.6 s & 11.7 min & 9.4 & $\sim$10.5 h \\
\textbf{CR-OPD}
& \textbf{37.7} & 871 & 6.0 s & 10.7 min & 7.5 & $\sim$10.5 h \\

\bottomrule
\end{tabular}%
}
\end{table*}
The unified latency experiment uses identical hardware, batch size, decoding
configuration, and warmed-up model execution so that model-loading and process
startup do not distort per-sample latency.
Once the timing measurements are completed, Table~\ref{tab:efficiency_full}
will allow recovery gains to be interpreted jointly with realized inference
and training cost.


\section{Qualitative Recovery and Regression Cases}
\label{app:case_studies}

Qualitative inspection complements the aggregate stratified results.
Recovered examples typically involve problems for which the base model
completes a substantial reasoning trajectory but makes an intermediate
algebraic, arithmetic, assumption, or verification error. CR-SFT often
produces a more complete derivation or introduces an additional check before
committing to the final answer.
Representative recovery cases are summarized in
Table~\ref{tab:recovery_cases}.
\begin{table*}[t]
\centering
\small
\caption{
Representative recovery cases.
Exact examples will be inserted after final manual selection.
}
\label{tab:recovery_cases}
\setlength{\tabcolsep}{8pt}
\renewcommand{\arraystretch}{1.12}

\begin{tabular*}{0.97\textwidth}{
@{\extracolsep{\fill}}
p{0.18\textwidth}
p{0.35\textwidth}
p{0.35\textwidth}
}
\toprule
\textbf{Problem}
& \textbf{Base failure}
& \textbf{CR-SFT recovery} \\
\midrule

Triangle $ABC$, $a{=}12,b{=}17,c{=}13$; compute
$b\cos C-c\cos B$. (Ans.\ 10)
&
Correct Law-of-Cosines setup but an arithmetic slip
($144{+}169$ taken as $213$ not $313$) yields
$\cos B=-\tfrac{19}{78}$ and the wrong answer
$\tfrac{1105}{78}$.
&
Same method, correct arithmetic:
$\cos B=\tfrac1{13}$, $\cos C=\tfrac{11}{17}$, so
$17\cdot\tfrac{11}{17}-13\cdot\tfrac1{13}
=11-1=\mathbf{10}$.
\\

\addlinespace[5pt]

Max.\ checkers on an $8\times8$ board so each is diagonally attacked.
(Ans.\ 32)
&
Finds a valid 32-checker pattern, then discards it and claims all
$64$ squares work (``every square is adjacent to others''), missing
the upper bound. Answers $\mathbf{64}$.
&
Keeps the 32-configuration and proves any 33rd checker must be left
isolated (unattacked); hence the maximum is $\mathbf{32}$.
\\

\addlinespace[5pt]

Rearrangements of \textsc{ccamb} with at least one C before the A.
(Ans.\ 40)
&
Reduces to the order of $\{A,C,C\}$ but miscounts all three orderings
(ACC, CAC, CCA) as ``C before A''---ACC is not---so concludes all
$60$ permutations qualify.
&
Only CAC and CCA qualify ($2$ of $3$), giving probability
$\tfrac23$, so $60\cdot\tfrac23=\mathbf{40}$.
\\

\bottomrule
\end{tabular*}
\end{table*}
We also inspect regressions on already-solved examples. These cases tend to be
problems that the base model originally solves quickly and correctly, but whose
reasoning is perturbed after post-training. Reporting both recovery and
regression examples is important because the aggregate improvement in
Table~\ref{tab:general_main} arises from a trade-off rather than uniform
monotonic improvement.

\section{Cross-Model Collection and Overlap}
\label{app:cross_model_details}

CR membership is defined by a model's observed response to additional
reasoning, so it need not transfer directly across models.
To test this, we reconstruct CR sets from the same underlying reasoning sources
using each model's own low-compute failures and additional-reasoning
trajectories.
This setup controls the source pool while allowing the selected CR examples to
vary by model.

\begin{figure}[t]
\centering
\includegraphics[width=\linewidth]{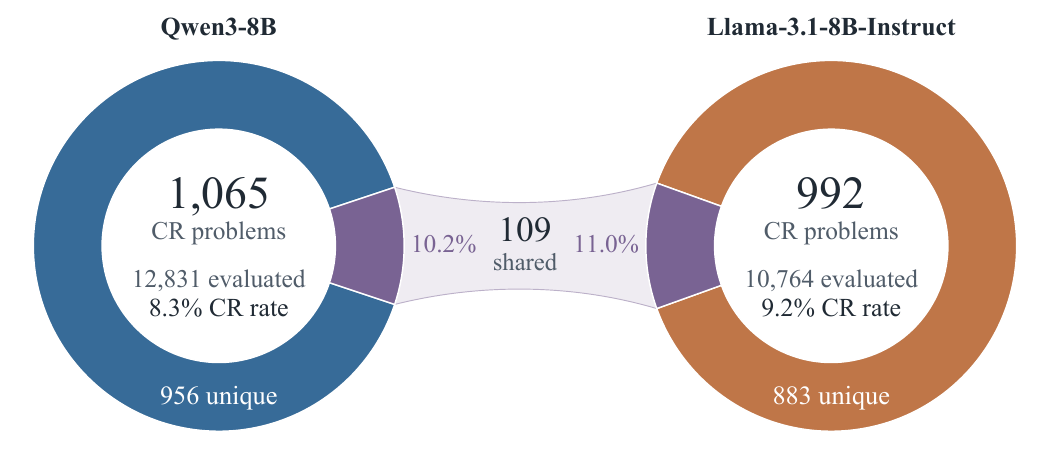}
\caption{
\textbf{Cross-model overlap of compute-responsive examples.}
Qwen3-8B and Llama-3.1-8B-Instruct identify 1,065 and 992 CR problems,
respectively, with only 109 examples shared between the two CR sets.
The shared examples account for 10.2\% of the Qwen3-8B CR set and 11.0\% of
the Llama-3.1-8B-Instruct CR set.
}
\label{fig:cross_model_overlap}
\end{figure}

Table~\ref{tab:cross_model_overlap} reports the resulting CR-set statistics,
while Figure~\ref{fig:cross_model_overlap} visualizes the overlap between the
two model-specific CR sets.
The overlap column measures how many CR examples for each model also appear in
the Qwen3-8B CR set.
A low overlap indicates that the model identifies a different subset of
recoverable failures from the same source pool, rather than simply recovering a
fixed set of intrinsically responsive problems.
\begin{table*}[t]
\centering
\small
\caption{
\textbf{Cross-model CR-set statistics.}
CR examples are reconstructed separately for each model from the same
underlying reasoning sources. The overlap column reports overlap with the
Qwen3-8B CR set.
}
\label{tab:cross_model_overlap}
\setlength{\tabcolsep}{10pt}
\renewcommand{\arraystretch}{1.14}

\begin{tabular*}{0.92\textwidth}{
@{\extracolsep{\fill}}
lrrrr
}
\toprule
Model
& Evaluated
& CR Count
& CR Rate
& \shortstack{Overlap with\\Qwen3-8B CR} \\
\midrule

Qwen3-8B
& 12,831
& 1,065
& 8.3\%
& -- \\

Llama-3.1-8B-Instruct
& 10,764
& 992
& 9.2\%
& 109 (11.0\%) \\

\bottomrule
\end{tabular*}
\end{table*}
The overlap is limited.
Llama-3.1-8B-Instruct identifies 992 CR examples, but only 109 of them overlap
with the Qwen3-8B CR set, corresponding to 11.0\% of the Llama CR set.
These results support the intended interpretation of CR membership as
model-dependent: the same source pool can contain different recoverable
boundaries for different models.

This analysis also serves as a fixed-source control.
If compute responsiveness were merely a property of a fixed subset of
questions, different models would be expected to recover largely the same CR
examples from the same pool.
Instead, the partial overlap suggests that CRDS is not selecting a universal
list of easy-to-recover questions.
Rather, it identifies the failures that are recoverable for the current model
under the specified additional-reasoning protocol.
The main cross-model training result in Table~\ref{tab:cross_model} then tests
the complementary question: whether each model's own CR set provides useful
post-training supervision.
\section{Training Stability}
\label{app:stability}
The primary experiments use frozen checkpoints selected on the validation set.
Because both SFT and OPD can exhibit training variance, we additionally repeat
the principal configurations with multiple training seeds where computational
budget permits.
The resulting stability statistics are reported in
Table~\ref{tab:stability}.
\begin{table}[t]
\centering
\small
\caption{
Multi-seed training stability.
}
\label{tab:stability}
\setlength{\tabcolsep}{8pt}
\renewcommand{\arraystretch}{1.12}

\begin{tabular*}{0.94\linewidth}{
@{\extracolsep{\fill}}
lrrrr
}
\toprule
Method
& Seed 1
& Seed 2
& Seed 3
& Mean $\pm$ Std. \\
\midrule

\textbf{SFT-CR}
& 46.2
& 41.5
& 49.1
& \textbf{45.6 $\pm$ 3.8} \\

SFT-Already-Solved
& 39.6
& 39.6
& 38.7
& 39.3 $\pm$ 0.5 \\

CR-OPD
& 33.0
& 32.1
& 34.9
& 33.3 $\pm$ 1.4 \\

\bottomrule
\end{tabular*}
\end{table}
Table~\ref{tab:stability} has been used to determine whether the observed differences are robust to training randomness rather than driven by a single seed.

\section{First-Hit Cost of CR Discovery}
\label{app:first_hit_cost}

\begin{figure}[t]
\centering
\includegraphics[width=\linewidth]{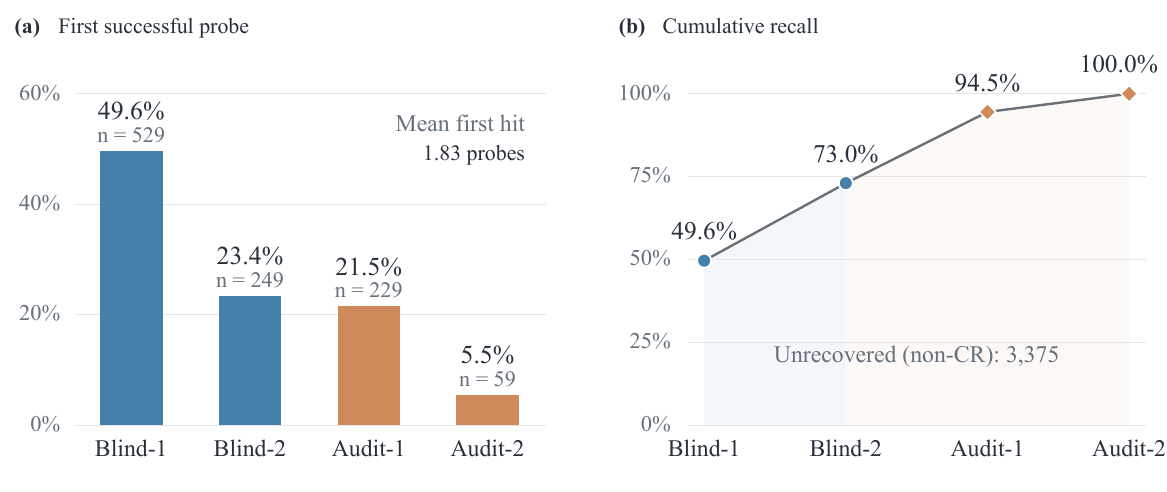}
\caption{
\textbf{First-hit distribution and cumulative recall during CR discovery.}
(a) Distribution of the first successful additional probe among CR examples.
(b) Cumulative recall after successive blind re-solving and audit-based
revision attempts.
Most CR examples are recovered within the first two blind probes, with a mean
first-hit cost of 1.83 additional generations.
}
\label{fig:first_hit_cost}
\end{figure}

Our CR construction protocol allows up to four additional trajectories for each
complete initial failure: two sampled blind re-solving attempts followed by two
audit-based revision attempts.
This design is intended to improve recall of recoverable failures, but it does
not mean that most CR examples require all four probes to be detected.
Table~\ref{tab:first_hit_cost} reports the first additional trajectory that
successfully recovers each CR example, while
Figure~\ref{fig:first_hit_cost} visualizes the first-hit distribution and
cumulative recall across probes.
\begin{table}[h]
\centering
\small
\caption{
\textbf{First successful additional trajectory among CR examples.}
Most CR examples are detected before all four probes are used.
}
\label{tab:first_hit_cost}
\setlength{\tabcolsep}{9pt}
\renewcommand{\arraystretch}{1.12}

\begin{tabular*}{0.94\linewidth}{
@{\extracolsep{\fill}}
lrrr
}
\toprule
First successful probe
& Count
& Share among CR
& Cumulative recall \\
\midrule

Blind-1
& 529 & 49.6 & 49.6 \\
Blind-2
& 249 & 23.4 & 73.0 \\
Audit-1
& 229 & 21.5 & 94.5 \\
Audit-2
& 59 & 5.5 & 100.0 \\

\addlinespace[2pt]
\midrule
Never recovered (non-CR)
& 3375
& --
& -- \\

\bottomrule
\end{tabular*}
\end{table}
A single blind re-solving attempt detects 49.6\% of all CR examples, and two
blind attempts detect 73.0\%.
The mean first-hit cost among CR examples is
\[
\frac{
1 \cdot 529 + 2 \cdot 249 + 3 \cdot 229 + 4 \cdot 59
}{
529 + 249 + 229 + 59
}
= 1.83
\]
additional generations.
Thus, the four-probe protocol should be viewed as a recall-oriented upper bound
rather than the typical cost required to identify a CR example.
The audit probes nevertheless provide non-redundant signal.
Among all CR examples, 288 examples are first recovered by audit-based revision
rather than blind re-solving, accounting for 27.0\% of the CR set.
This shows that audit trajectories recover a substantial subset of failures not
found by blind resampling alone, while the marginal gain from the fourth probe
is small.

\section{Retention--Recovery Trade-off under Mixed Training}
\label{app:mix_training}

The main experiments use matched single-source training sets to isolate the
effect of compute-responsive selection.
This design is not intended to optimize the final deployment mixture.
In practice, one can combine CR examples with examples that preserve behavior
on problems the base model already solves.
We therefore evaluate a simple 50\% CR + 50\% Already-Solved mixture as a
deployment-oriented trade-off analysis for SFT.

Table~\ref{tab:mix_training} compares CR-only, Already-Solved-only, and mixed
SFT on the mixed generalization set.
SFT-Mix improves accuracy on the already-solved stratum to 87.4\%, compared
with 86.3\% for SFT-CR and 87.1\% for SFT-Already-Solved.
Although CR-stratum accuracy decreases from 47.2\% under SFT-CR to 43.4\%
under SFT-Mix, it remains substantially higher than the 30.2\% achieved by
SFT-Already-Solved.
The mixture also preserves part of the improvement on unresolved examples,
reaching 3.5\%, compared with 2.7\% for SFT-Already-Solved and 3.9\% for
SFT-CR.
These results show that the regression on already-solved examples can be
mitigated by adjusting the post-training mixture, without discarding the
recovery benefit of CR examples.

\begin{table}[t]
\centering
\small
\caption{
\textbf{Retention--recovery trade-off under mixed SFT.}
Mixed training combines 50\% CR examples with 50\% Already-Solved examples.
}
\label{tab:mix_training}
\setlength{\tabcolsep}{10pt}
\renewcommand{\arraystretch}{1.12}

\begin{tabular*}{0.90\linewidth}{
@{\extracolsep{\fill}}
lrrr
}
\toprule
Method
& Already Solved
& CR
& Unresolved \\
\midrule

SFT-Already-Solved
& 87.1
& 30.2
& 2.7 \\

SFT-CR
& 86.3
& \textbf{47.2}
& \textbf{3.9} \\

\textbf{SFT-Mix}
& \textbf{87.4}
& 43.4
& 3.5 \\

\bottomrule
\end{tabular*}
\end{table}
\section{Teacher--Student KL Diagnostic in OPD}
\label{app:opd_kl_diagnostic}

The OPD experiments use Qwen3-8B as the student and Qwen3-32B as the frozen
teacher.
The training dynamics provide a diagnostic view of why CR-selected states are
useful under OPD.
For each training condition, we measure the teacher--student KL divergence on
student on-policy states and compare the average KL at the beginning and end of
training.
Table~\ref{tab:opd_kl_diagnostic} reports the mean KL, the average KL over the
first 10 training steps, the average KL over the final 10 steps, and the
qualitative trajectory.

\begin{table}[t]
\centering
\small
\caption{
\textbf{Teacher--student KL dynamics under OPD.}
CR-selected states exhibit the largest teacher--student divergence and the
clearest decrease during training, indicating a stronger and more learnable
distillation signal.
}
\label{tab:opd_kl_diagnostic}
\setlength{\tabcolsep}{7pt}
\renewcommand{\arraystretch}{1.12}

\begin{tabular*}{\linewidth}{
@{\extracolsep{\fill}}
lrrrl
}
\toprule
Condition
& Mean KL
& First 10 steps
& Last 10 steps
& Trajectory \\
\midrule

\textbf{OPD-CR}
& \textbf{0.472}
& \textbf{0.564}
& 0.420
& \textbf{Clear decrease} \\

OPD-All
& 0.366
& 0.389
& 0.344
& Mild decrease \\

OPD-AlreadySolved
& 0.354
& 0.346
& 0.346
& Flat \\

\bottomrule
\end{tabular*}
\end{table}

The KL pattern supports the interpretation that CR examples expose states where
the teacher provides more informative supervision.
OPD-CR has the largest mean KL, 0.472, compared with 0.366 for OPD-All and
0.354 for OPD-AlreadySolved.
The difference is especially pronounced early in training: the first-10-step KL
is 0.564 for OPD-CR, 0.389 for OPD-All, and 0.346 for OPD-AlreadySolved.
Thus, on CR-selected states, the teacher and student initially disagree more
strongly than on already-solved states.
The trajectory of the KL divergence is also informative.
For OPD-CR, KL decreases from 0.564 in the first 10 steps to 0.420 in the last
10 steps, indicating that the student absorbs part of the teacher distribution
on states where it originally fails.
OPD-All shows a smaller decrease, from 0.389 to 0.344.
OPD-AlreadySolved remains essentially flat at 0.346, suggesting that the
student and teacher are already closely aligned on these states and that the
available gradient signal is limited.

This diagnostic explains why a stronger teacher is useful in the CR setting.
CR examples are selected precisely because the 8B student fails initially while
the problem remains recoverable under additional reasoning.
Such states expose larger teacher--student disagreements than already-solved
examples, and OPD converts this disagreement into a training signal.
When the student is already aligned with the teacher, as in the
Already-Solved condition, the KL remains low and flat, leaving little additional
supervision to absorb.

\section{Construction of the Mixed Generalization Test Set}
\label{app:general_test_construction}

The held-out CR test split contains 106 compute-responsive examples and is used
to measure recovery on problems that are complete initial failures for the base
model.
To evaluate whether CR training affects a broader distribution, we construct an
additional mixed generalization test set by combining this frozen CR test split
with independently sampled Already-Solved and Unresolved examples from the same
source pools.

For each source dataset, we first form two non-CR pools using the base model's
collection-time behavior.
The Already-Solved pool contains problems solved by the initial low-budget pass.
The Unresolved pool contains complete low-budget failures that are not recovered
by the additional-reasoning probes.
We then sample non-CR examples from each dataset in proportion to the natural
Already-Solved/Unresolved composition of that dataset.
For a dataset with target sample size $m_d$, the allocated counts are
\begin{equation}
n^{(d)}_{\mathrm{AS}}
=
\mathrm{round}
\left(
m_d
\frac{
N^{(d)}_{\mathrm{AS}}
}{
N^{(d)}_{\mathrm{AS}} + N^{(d)}_{\mathrm{HL}}
}
\right),
\qquad
n^{(d)}_{\mathrm{HL}}
=
m_d - n^{(d)}_{\mathrm{AS}},
\end{equation}
where $N^{(d)}_{\mathrm{AS}}$ and $N^{(d)}_{\mathrm{HL}}$ denote the available
Already-Solved and Unresolved pools for dataset $d$.
We use $m_d=150$ when enough non-CR examples are available; for AIME, where the
non-CR pool contains only 29 examples, we include all available examples.

Table~\ref{tab:general_test_construction} reports the resulting allocation.
The sampled non-CR portion contains 372 Already-Solved and 257 Unresolved
examples.
Combining these with the 106-example frozen CR test split gives the
735-example mixed generalization test set used in
Table~\ref{tab:general_main}.

\begin{table}[t]
\centering
\small
\caption{
\textbf{Construction of the non-CR portion of the mixed generalization test set.}
Already-Solved and Unresolved examples are sampled within each source dataset
according to their natural proportions in the available non-CR pool.
}
\label{tab:general_test_construction}
\setlength{\tabcolsep}{5.5pt}
\renewcommand{\arraystretch}{1.10}

\begin{tabular*}{\linewidth}{
@{\extracolsep{\fill}}
lrrrr
}
\toprule
&
\multicolumn{2}{c}{\textbf{Available non-CR pool}}
&
\multicolumn{2}{c}{\textbf{Sampled test set}}
\\
\cmidrule(lr){2-3}
\cmidrule(lr){4-5}

Dataset
& Already-Solved
& Unresolved
& Already-Solved
& Unresolved \\
\midrule

AIME
& 13 & 16 & 13 & 16 \\

DeepScaleR
& 1243 & 737 & 94 & 56 \\

NuminaMath-CoT
& 2273 & 1415 & 92 & 58 \\

OlympiadBench
& 254 & 182 & 87 & 63 \\

OpenThoughts
& 762 & 566 & 86 & 64 \\

\midrule
\textbf{Total non-CR}
& --
& --
& \textbf{372}
& \textbf{257} \\

\addlinespace[2pt]
Frozen CR test split
& --
& --
& \multicolumn{2}{c}{\textbf{106}} \\

\midrule
\textbf{Mixed test set total}
& --
& --
& \multicolumn{2}{c}{\textbf{735}} \\

\bottomrule
\end{tabular*}
\end{table}
\end{document}